\documentclass[10pt,letterpaper]{article}

\usepackage[round]{natbib}
\usepackage{bibunits}

\usepackage[T1]{fontenc}
\usepackage[utf8]{inputenc}
\usepackage[english]{babel}

\usepackage{amsmath,amssymb,amsthm}
\usepackage{XCharter}
\usepackage[charter]{newtxmath}
\IfFileExists{sourcesans.sty}{%
  \usepackage[scale=0.95]{sourcesans}%
}{%
  \usepackage[scale=0.95]{sourcesanspro}%
}
\IfFileExists{zi4.sty}{%
  \usepackage[scaled=0.95]{zi4}%
}{}

\usepackage{mathtools,bm}
\usepackage{microtype}

\usepackage{graphicx}
\usepackage{booktabs,array,tabularx,multirow}
\usepackage[table]{xcolor}
\usepackage[ruled,vlined,noend]{algorithm2e}
\usepackage{enumitem}
\setlist[itemize]{leftmargin=*,itemsep=2pt,topsep=4pt,parsep=0pt,partopsep=0pt}
\setlist[enumerate]{leftmargin=*,itemsep=2pt,topsep=4pt,parsep=0pt,partopsep=0pt}

\definecolor{ink}{HTML}{1E242B}          
\definecolor{physics}{HTML}{0F4D92}      
\definecolor{neural}{HTML}{3775BA}       
\definecolor{feedback}{HTML}{B64342}     
\definecolor{layoutmuted}{HTML}{5C6672}  

\newcommand{\ourmethod}{{\normalfont\textsc{Forge}}}
\renewcommand{\eqref}[1]{\textup{(\ref{#1})}}
\newcommand{\best}[1]{\cellcolor{physics!12}{\boldmath\textbf{#1}}}
\newcommand{\second}[1]{\underline{#1}}
\newcommand{\sd}[1]{{\unboldmath\normalfont\fontsize{6.5}{8}\selectfont\textcolor{ink!62}{$\,\pm\,#1$}}}
\newcommand{\R}{\mathbb{R}}

\newcommand{\dd}{\mathrm{d}}
\newtheorem{proposition}{Proposition}

\newenvironment{wideblock}{\par\vspace{6pt}}{\par\vspace{6pt}}

\newcommand{\figfile}[2][]{%
  \IfFileExists{fig/#2}{\includegraphics[#1]{#2}}{%
    \fbox{\parbox{0.92\linewidth}{\centering\vspace{18pt}\texttt{\detokenize{#2}}\vspace{18pt}}}}}

\usepackage[textwidth=166mm,top=24mm,bottom=24mm,headheight=14pt,headsep=15pt,footskip=25pt]{geometry}
\usepackage{fancyhdr,titlesec,caption}

\makeatletter
\renewcommand{\normalsize}{%
  \@setfontsize\normalsize{10.5}{14.1}%
  \abovedisplayskip=8pt plus 2pt minus 2pt
  \belowdisplayskip=8pt plus 2pt minus 2pt
  \abovedisplayshortskip=3pt plus 1pt
  \belowdisplayshortskip=5pt plus 1pt}
\renewcommand{\small}{\@setfontsize\small{9.3}{12.0}}
\renewcommand{\footnotesize}{\@setfontsize\footnotesize{8.5}{10.8}}
\makeatother
\normalsize

\titleformat{\section}{\sffamily\bfseries\fontsize{14.5}{18}\selectfont\color{physics}}{\thesection}{0.8em}{}
\titleformat{\subsection}{\sffamily\bfseries\fontsize{11.8}{14.5}\selectfont\color{ink}}{\thesubsection}{0.65em}{}
\titleformat{\subsubsection}{\sffamily\bfseries\normalsize\color{ink}}{\thesubsubsection}{0.65em}{}
\titleformat{\paragraph}[runin]{\bfseries\normalsize\color{ink}}{\theparagraph}{0.5em}{}
\titlespacing*{\section}{0pt}{19pt plus 3pt minus 2pt}{8pt}
\titlespacing*{\subsection}{0pt}{14pt plus 2pt minus 2pt}{6pt}
\titlespacing*{\subsubsection}{0pt}{10pt plus 2pt}{5pt}
\titlespacing*{\paragraph}{0pt}{8pt plus 1pt}{0.8em}

\DeclareCaptionFont{forgecaption}{\fontsize{9.2}{12.2}\selectfont}
\renewenvironment{abstract}{%
  \par\begingroup\fontsize{9.7}{13.1}\selectfont\setlength{\parskip}{5pt}%
  \noindent{\sffamily\bfseries\fontsize{10}{13}\selectfont\color{physics}Abstract}\par\vspace{4pt}%
}{\par\endgroup\vspace{14pt}}

\newcommand{\shorttitle}{Sounding the Black Box}
\fancypagestyle{plain}{%
  \fancyhf{}%
  \fancyfoot[R]{\sffamily\fontsize{8.5}{10}\selectfont\color{layoutmuted}\thepage}%
  }

\usepackage{hyperref}
\usepackage{url}
\hypersetup{
  colorlinks=true,
  linkcolor=ink,
  citecolor=physics,
  urlcolor=physics,
  pdfborder={0 0 0}
}

\newcommand{\PaperLeadTitle}{}
\newcommand{\PaperSubTitle}{}
\newcommand{\MakeForgeTitle}{%
  \thispagestyle{plain}%
  {\sffamily\bfseries\fontsize{9}{11}\selectfont\color{physics}\ourmethod\par}%
  \vspace{14pt}%
  {\fontsize{31}{35}\selectfont\bfseries\color{ink}\PaperLeadTitle\par}%
  \ifx\PaperSubTitle\empty\else
    \vspace{7pt}%
    {\fontsize{18.5}{23.5}\selectfont\raggedright\color{ink}\PaperSubTitle\par}%
  \fi
  \vspace{16pt}%
  {\fontsize{10.8}{14}\selectfont\raggedright\color{ink}\AuthorList\par}%
  \vspace{7pt}%
  {\fontsize{8.8}{11.5}\selectfont\raggedright\color{layoutmuted}\AffiliationList\par}%
  \vspace{5pt}%
  {\fontsize{8.4}{11}\selectfont\raggedright\color{layoutmuted}\AuthorNotes\par}%
  \vspace{15pt}%
  {\color{physics}\hrule height 0.65pt}%
  \vspace{16pt}%
}

\newcommand{\AuthorList}{%
  Ruiqi Shu\textsuperscript{1,*}\quad
  Tianrun Gao\textsuperscript{2,*}\quad
  Haomin Wen\textsuperscript{3,\ensuremath{\dagger}}\quad
  Xiaomeng Huang\textsuperscript{1,\ensuremath{\dagger}}%
}

\newcommand{\AffiliationList}{%
  \textsuperscript{1}\,Department of Earth System Science, Tsinghua University, Beijing, China\\
  \textsuperscript{2}\,School of Computing and Data Science, The University of Hong Kong, Hong Kong, China\\
  \textsuperscript{3}\,Shanghai Innovation Institute, Shanghai, China%
}

\newcommand{\AuthorNotes}{%
  \textsuperscript{*}\,Equal contribution.\quad
  \textsuperscript{\ensuremath{\dagger}}\,Corresponding authors:\\
  Haomin Wen (\href{mailto:wenhaomin.whm@gmail.com}{\texttt{wenhaomin.whm@gmail.com}}),\quad
  Xiaomeng Huang (\href{mailto:hxm@tsinghua.edu.cn}{\texttt{hxm@tsinghua.edu.cn}}).%
}

\graphicspath{{fig/}}

\renewcommand{\PaperLeadTitle}{Sounding the Black Box}
\renewcommand{\PaperSubTitle}{Online Training of Embedded Neural Networks in Forward-Only Physical Simulators}
\renewcommand{\shorttitle}{Sounding the Black Box: Online Training in Forward-Only Physical Simulators}

\hypersetup{
  pdftitle={Sounding the Black Box: Online Training of Embedded Neural Networks in Forward-Only Physical Simulators},
  pdfauthor={Ruiqi Shu; Tianrun Gao; Haomin Wen; Xiaomeng Huang}
}

\begin{document}

\MakeForgeTitle

\vspace{-8pt}
\begin{center}
    \captionsetup{type=figure}
    \figfile[width=\textwidth]{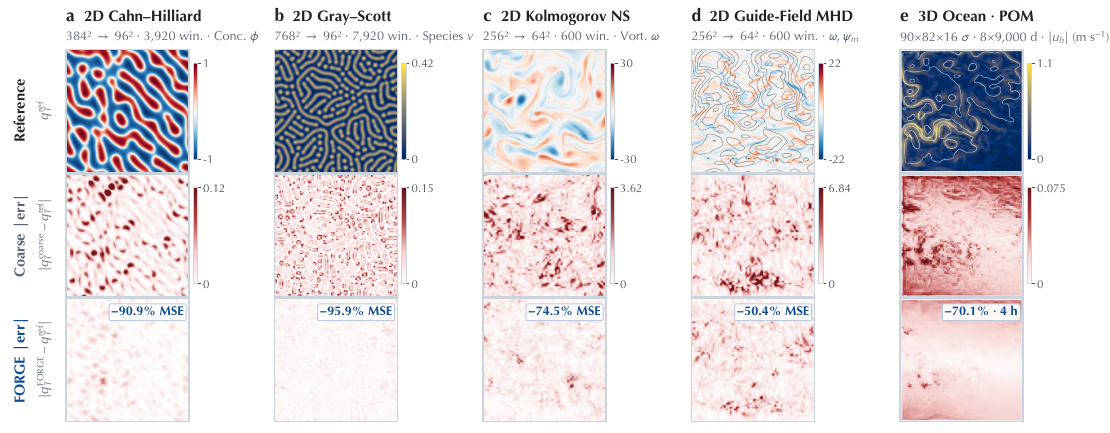}\vspace{3pt}
    \caption{\textbf{Closed-loop multi-physics benchmark and rollout error reductions.}
    \textbf{Top row:} reference fields $q_T^{\mathrm{ref}}$ at the final step $T$ of a validation window across all five PDE regimes (\textbf{a}--\textbf{e}: concentration $\phi$, reactant $v$, vorticity $\omega$, $\omega$ with magnetic potential $\psi_m$ contours, and ocean surface speed).
    \textbf{Middle and bottom rows:} pointwise absolute errors ($|\mathrm{err}|$) for the unparameterized coarse solver ($G_\theta \equiv 0$) and \ourmethod{} under a shared error scale per column.}
    \label{fig:benchmarks}
\end{center}
\vspace{4pt}

\begin{bibunit}[plainnat]
\begin{abstract}
Hybrid physics--neural simulators embed neural networks in partial differential equation (PDE) solvers to model unresolved subgrid dynamics while preserving resolved physics and conservation laws. Because closure outputs feed back across time steps, single-step offline regression does not directly minimize multi-step rollout error, motivating online trajectory optimization. Solver backpropagation through time (BPTT) is the standard online solution when solver derivatives are available, yet many operational Fortran and C/C++ simulators expose only non-differentiable, forward-only interfaces. We introduce \ourmethod{} (\textbf{F}orward-\textbf{O}nly \textbf{R}esponse and \textbf{G}auss--Newton \textbf{E}stimation), an online framework that trains embedded networks from forward runs alone by decoupling exact network derivatives from black-box simulator dynamics. Stage~I warms up the rollout via a low-dimensional Gauss--Newton step along output-orthonormal weight directions, canceling multi-layer feature drift via a final-layer correction. Stage~II perturbs spatial output fields to fit a gain-bounded causal linear response operator across rollout steps, solves for regularized target outputs, and updates all weights via network backpropagation. Across five 2D and 3D benchmarks spanning phase separation, reaction--diffusion, fluid and MHD turbulence, and ocean circulation, \ourmethod{} outperforms gradient-free, reinforcement-learning, and emulator-based baselines, cutting coarse-solver test mean squared error (MSE) by $50\%$--$96\%$ in 2D and $70\%$ in 3D, outperforming the strongest forward-only baseline by $16\%$--$69\%$, and recovering $88\%$--$98\%$ of Solver BPTT's gains. Code is available at \url{https://github.com/ChiyodaMomo01/FORGE}.
\end{abstract}

\section{Introduction}
\label{sec:intro}

\begin{figure*}[t]
  \centering
  \figfile[width=0.95\textwidth]{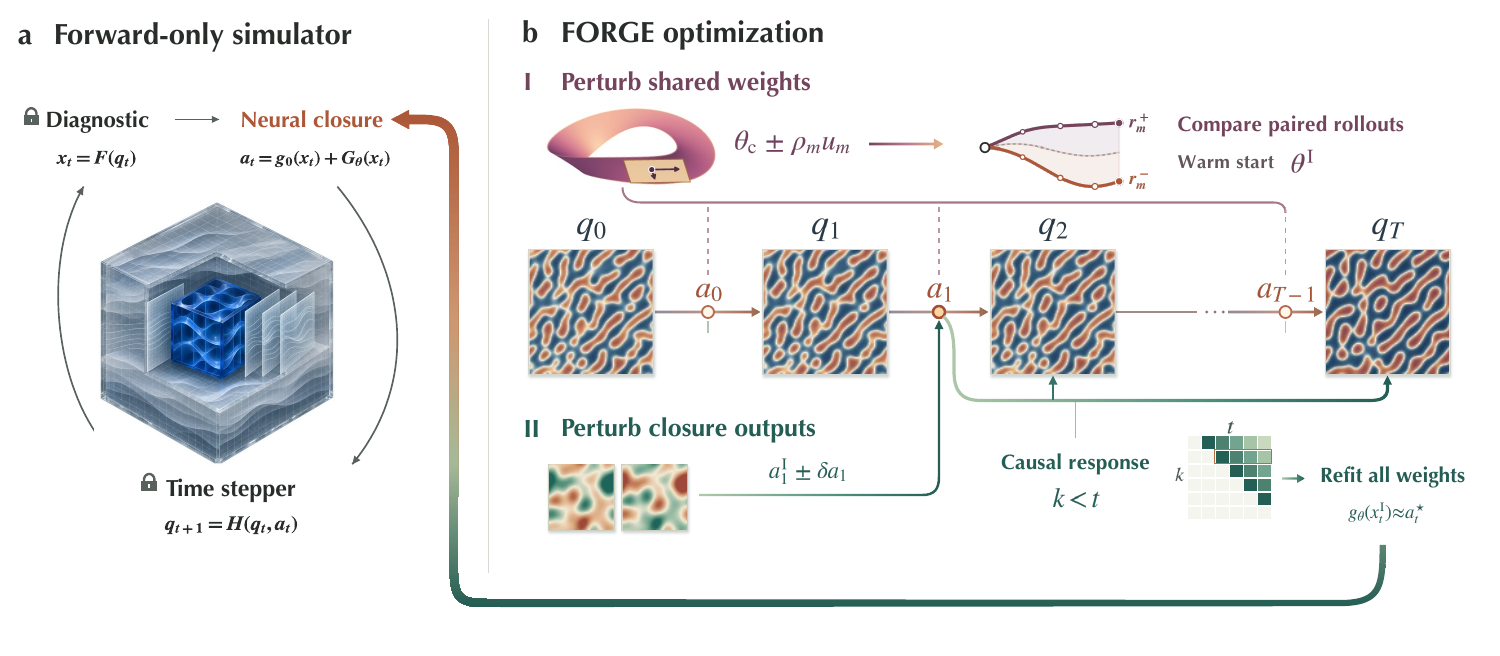}
  \caption{\textbf{Problem overview and two-stage \ourmethod{} framework.} \textbf{a}, \textbf{Forward-only hybrid simulation:} a neural closure $G_\theta$ is embedded inside a coarse PDE solver ($F \to g_\theta \to H$) that returns forward state trajectories $q_0 \to \dots \to q_T$ without solver derivatives. \textbf{b}, \textbf{Two-stage online training:} \textbf{Stage~I} perturbs weights along output-orthonormal directions with readout drift compensation to warm up the rollout to $\theta^{\mathrm{I}}$; \textbf{Stage~II} perturbs closure outputs around $\theta^{\mathrm{I}}$ to fit a causal response operator $\mathcal{R}$, solves for target outputs $a_t^\star$, and updates all weights via network backpropagation.}
  \label{fig:overview}
\end{figure*}

Simulating multi-scale partial differential equations (PDEs) across physics-grounded visual simulation, reactive patterns, turbulence, and Earth-system forecasting is computationally prohibitive at full resolution. Standalone neural surrogates replace numerical time integration without enforcing discrete conservation updates \citep{li2021fourier,takamoto2022pdebench,cui2025forecasting}. \emph{Hybrid physics--neural simulators} instead retain a coarse numerical PDE solver and embed a \emph{neural closure} for unresolved subgrid fluxes or sources, inheriting resolved transport, conservation laws, and boundary conditions \citep{rasp2018deep,um2020solver,kochkov2021machine,kochkov2024neural}. Training an embedded closure, however, is a coupled multi-step problem: each closure output alters subsequent solver states across the rollout. Although carefully designed offline closures can integrate stably \citep{rasp2018deep,yuval2020stable,perezhogin2024stable}, single-step snapshot regression does not directly minimize multi-step trajectory error and often drifts under closed-loop feedback \citep{brenowitz2018prognostic,ross2023benchmarking}. When solver derivatives are available, \emph{solver backpropagation through time} (Solver BPTT) \citep{werbos1990backpropagation,um2020solver,frezat2022posteriori,shu2026hybridom} unrolls the trajectory and differentiates the multi-step loss through every solver step. Yet Solver BPTT requires maintained adjoints or differentiable rewrites \citep{bezgin2023jaxfluids,shen2023differentiable}, whereas many operational Fortran and C/C++ simulators expose only forward-only interfaces without computational graphs or adjoints (Fig.~\ref{fig:overview}a).

This raises a core question: \emph{when a simulator exposes only forward evaluations without derivatives, how can we train an embedded neural network online and approach the closed-loop accuracy of Solver BPTT?} Existing forward-only methods face three bottlenecks:
(1) \emph{Zeroth-order and ensemble methods} (ES \citep{salimans2017evolution}, subspace Gauss--Newton \citep{cartis2023scalable}, and EKI \citep{kovachki2019ensemble,lopezgomez2022training}) perturb weights directly, treating the rollout as a static black box; in deep closures, small budgets span tiny subspaces where random weight steps barely move outputs, while finite multi-layer steps distort tangents.
(2) \emph{Reinforcement learning methods} (PPO, multi-agent turbulence RL, and Closure-RL \citep{schulman2017proximal,novati2021automating,bassewitz2025closure}) inject step-wise action noise and optimize scalar, spectral, or grid-local rewards without cross-step state Jacobians.
(3) \emph{Emulator-based and single-step approximations}: the \emph{Neural Emulator Method} \citep{wright2022deep,frezat2026gradient} fits an auxiliary solver emulator online to backpropagate multi-step updates into the embedded closure (still running the true coarse solver at test time), yet emulator state and gradient errors compound across steps; static single-step sensitivities (cf.\ \citet{ouala2024online}) miss multi-step error propagation and spatial or implicit solver operators.

To overcome these bottlenecks, we propose \ourmethod{} (Fig.~\ref{fig:overview}b), a two-stage online framework that decouples exact neural network derivatives from black-box simulator dynamics and \textbf{exploits the causal temporal structure of time-stepped rollouts}. Rather than collapsing a $T$-step rollout into a static loss, \ourmethod{} tracks how closure shifts at step $k$ propagate into state errors at steps $t > k$. In each round, \textbf{Stage~I} uses forward-mode network derivatives to probe a low-dimensional subspace of output-orthonormal weight directions with final-layer drift compensation, warming up the rollout into a locally linear regime. Around this trajectory, \textbf{Stage~II} perturbs spatial closure outputs to fit a structured \textbf{causal error-propagation operator} ($d_{t+1} = A_t d_t + B_t\,\delta a_t$) on the solver grid, solves for regularized step-wise target outputs, and updates all weights via network backpropagation without extra solver calls. Our main contributions are:
\begin{itemize}[leftmargin=1.3em,itemsep=1pt,topsep=1.5pt]
  \item \textbf{Multi-physics forward-only hybrid PDE benchmark:} Extending PDE closure protocols \citep{takamoto2022pdebench,ross2023benchmarking}, we build five coupled 2D and 3D benchmarks across phase separation, reaction--diffusion, fluid and MHD turbulence, and a research-grade 3D primitive-equation ocean circulation model \citep{blumberg1987description,shu2026neuralpom} under forward-only interfaces, evaluated across six forward-only baseline families and Solver BPTT references.
  \item \textbf{Two-stage forward-only online training (\ourmethod{}):} We propose a black-box trajectory optimization framework pairing output-orthonormal weight-subspace warm-up and readout drift compensation (Stage~I) with causal multi-step output-response identification and decoupled full-network backpropagation (Stage~II).
  \item \textbf{Theoretical guarantees and closed-loop validation:} We prove nonlinear residual cancellation in Stage~I and regularized target stability with error-tolerant descent in Stage~II. Across all five benchmarks, \ourmethod{} cuts coarse test mean squared error (MSE) by $50\%$--$96\%$ in 2D and $70\%$ in 3D, outperforms the strongest forward-only baseline by $16\%$--$69\%$, matches its final accuracy using $42\%$--$80\%$ fewer rollouts, recovers $88\%$--$98\%$ of Solver BPTT's gains, and stays stable up to $8\times$--$12\times$ the training horizon.
\end{itemize}

\section{Related Work}
\label{sec:related}

\paragraph{Hybrid PDE closures and differentiable simulation.}
Embedded neural closures parameterize unresolved fluxes and tendencies inside coarse PDE integrators \citep{ling2016reynolds,barsinai2019learning,duraisamy2019turbulence,foxkemper2019challenges,hewitt2020resolving,sirignano2020dpm,yin2021augmenting}, whereas pure neural surrogates replace time integration directly \citep{li2021fourier,takamoto2022pdebench,cui2025forecasting}. Offline \emph{a priori} snapshot training can yield stable climate or ocean runs under physical constraints or stochastic formulations \citep{rasp2018deep,bolton2019applications,zanna2020data,yuval2020stable,guillaumin2021stochastic,perezhogin2024stable}, yet single-step losses do not optimize multi-step accuracy and often drift under solver feedback \citep{brenowitz2018prognostic,beck2019deep,maulik2019subgrid,ross2023benchmarking}. Online \emph{a posteriori} Solver BPTT \citep{rumelhart1986learning,werbos1990backpropagation,rackauckas2020universal} differentiates unrolled trajectories or discrete adjoints across visual physics \citep{belbuteperes2018end,hu2020difftaichi,holl2020learning}, unsteady fluid and reacting flows \citep{um2020solver,kochkov2021machine,macart2021embedded,list2022learned,dresdner2023learning}, molecular dynamics and wavefields \citep{schoenholz2020jax,zhu2021general}, hydrology \citep{feng2022differentiable}, and atmosphere--ocean circulation \citep{frezat2022posteriori,kochkov2024neural,yan2025adjoint,shu2026hybridom,shu2026neuralpom}. Other differentiable pipelines bypass trajectory unrolling via implicit differentiation of steady or finite-element residuals \citep{belbuteperes2020combining,xue2023jaxfem}, Boltzmann reweighting for equilibrium observables \citep{thaler2021learning}, or surrogate calibration \citep{tsai2021calibration}. However, differentiable frameworks require solver rewrites or adjoints \citep{bezgin2023jaxfluids,gelbrecht2023differentiable,shen2023differentiable} and risk chaotic gradient growth \citep{list2025differentiability}, motivating online learning behind forward-only solver interfaces.

\paragraph{Derivative-free and ensemble parameter optimization.}
Zeroth-order estimators probe random weight perturbations or control variates \citep{spall1992multivariate,nesterov2017random,salimans2017evolution,grathwohl2018backpropagation}, derivative-free Gauss--Newton builds quadratic models in random weight subspaces \citep{cartis2023scalable}, and EKI calibrates climate parameters and neural closures via empirical covariances \citep{iglesias2013ensemble,kovachki2019ensemble,dunbar2021calibration,lopezgomez2022training}. Treating $\theta \mapsto r(\theta)$ as an unstructured map on $\R^p$ leaves unused both automatic differentiation of the neural network and the sequential step structure of the rollout.

\paragraph{Policy gradients, solver emulators, and trajectory response models.}
Reinforcement-learning closures apply PPO or multi-agent policy gradients using scalar, spectral, wall-stress, or grid-local rewards \citep{schulman2017proximal,novati2021automating,bae2022scientific,kurz2023deep,bassewitz2025closure}. Approximate-gradient methods replace solver differentiation with a learned backward pass \citep{wright2022deep}, layer target propagation \citep{lee2015difference}, the online Neural Emulator Method \citep{frezat2026gradient}, or online Euler gradient approximations (EGA) using static, tangent-linear, or ensemble state-propagation Jacobians for additive corrections \citep{ouala2024online}. Related assimilation and control methods either require solver adjoints, as in 4D-Var \citep{courtier1994strategy} and field inversion \citep{parish2016paradigm,duraisamy2019turbulence}, perturb initial states rather than step-wise closures in 4DEnVar \citep{liu2008ensemble,bannister2017review}, or fit unstructured low-dimensional transition matrices in Guided Policy Search \citep{levine2013guided,levine2014learning} (local, time-varying linear models in \citet{levine2014learning}) and DMDc \citep{proctor2016dynamic} that are underdetermined on PDE grids.

\section{Method}
\label{sec:method}

\subsection{Problem Formulation}
\label{sec:method-decomp}

A physical system $q(\mathbf{x}, t)$ evolves in continuous space and time at infinite resolution under a partial differential equation
\begin{equation}
\partial_t q = \mathcal{M}(q).
\label{eq:continuous-pde}
\end{equation}
Due to computational limits, numerical models can only evolve a discrete state $q_t$ on a low-resolution grid, advancing $q_t \mapsto q_{t+1}$ step by step. Running on a coarse grid leaves behind \emph{missing physics}, such as subgrid closures, unresolved reaction rates, or interfacial forces, causing the coarse solver alone to drift away from the reference states $q_t^{\mathrm{ref}}$.

A hybrid model retains the coarse numerical solver for resolved physics and conservation laws while embedding a neural network $G_\theta$ with parameters $\theta \in \R^p$ inside each step to represent this missing physics (Fig.~\ref{fig:overview}a). Because numerical time integration executes sequentially as a feed-forward pipeline, any solver step with an embedded closure decomposes into a stream $F \to g_\theta \to H$ of numerical stages over $t = 0, \dots, T-1$:
\begin{equation}
\begin{aligned}
x_t &= F(q_t), \quad a_t = g_\theta(x_t) = g_0(x_t) + G_\theta(x_t), \\
q_{t+1} &= H(q_t, a_t),
\end{aligned}
\label{eq:solver-decomp}
\end{equation}
where $F$ groups the pre-closure solver computations producing intermediate fields $x_t$ (e.g., spatial gradients, velocities, or predictor increments), $g_\theta$ adds the neural closure $G_\theta(x_t)$ to any baseline parameterization $g_0(x_t)$ ($g_0 \equiv 0$ for direct prediction), and $H$ groups the post-closure solver computations advancing the state (e.g., flux divergences, elliptic projections, or implicit solves), repeating $F \to g_\theta \to H$ across substeps when present. Stacking the normalized state errors $r_t(\theta) = q_t(\theta) - q_t^{\mathrm{ref}}$ over $t = 1, \dots, T$ into $r(\theta) \in \R^N$, training minimizes the closed-loop rollout MSE
\begin{equation}
\mathcal{L}(\theta) = \frac{1}{N}\|r(\theta)\|_2^2.
\label{eq:window-objective}
\end{equation}
When the solver is differentiable, \emph{Solver BPTT} \citep{werbos1990backpropagation,um2020solver,kochkov2024neural} computes the exact total derivative of \eqref{eq:window-objective} by applying the chain rule across all unrolled steps:
\begin{equation}
\frac{\dd\mathcal{L}(\theta)}{\dd\theta}
= \frac{2}{N}\sum_{t=1}^{T}\sum_{k=0}^{t-1}
\left(\frac{\partial G_\theta(x_k)}{\partial\theta}\right)^{\!\top}\!
\left(\frac{\dd q_t}{\dd a_k}\right)^{\!\top} r_t(\theta),
\label{eq:bptt-gradient}
\end{equation}
where $\frac{\dd q_t}{\dd a_k} = \frac{\dd q_t}{\dd q_{k+1}}\frac{\partial H(q_k, a_k)}{\partial a_k}$ propagates $a_k$ through $H$ and subsequent $(F, H)$ steps. Yet many operational Fortran and C/C++ simulators execute as non-differentiable, forward-only programs whose deployment interfaces expose state evaluations without adjoints or computational graphs, and our benchmarks likewise run each physical solver in an external process or with autograd disabled: while $\frac{\partial G_\theta}{\partial\theta}$ is given by automatic differentiation, the solver derivatives $\frac{\partial H}{\partial a_k}$, $\frac{\partial H}{\partial q}$, and $\frac{\partial F}{\partial q}$ inside $\frac{\dd q_t}{\dd a_k}$ are unavailable, precluding Solver BPTT. Instead, starting from $\theta_0$, the learner can only reset the simulator to $q_0$, evaluate $x_t \mapsto a_t$ forward, and observe $q_1(\theta), \dots, q_T(\theta)$. Below we split the neural network parameters $\theta = (h, b)$ into linear readout weights $h$ and hidden backbone weights $b$, writing $G_\theta(x) = \Phi(b; x)\,h$ with hidden features $\Phi(b; x)$.

\subsection{Motivation of Our Two-Stage Design}
\label{sec:method-motivation}

Stacking interval-held closure outputs $a = (a_0, \dots, a_{T-1})$ across the window, the cross-step solver sensitivities $\mathcal{J}_{a,tk} = \frac{\dd q_t}{\dd a_k}$ ($k < t$) in \eqref{eq:bptt-gradient} summarize the closed-loop trajectory Jacobian $\mathcal{J}_a = \frac{\dd r}{\dd a}$ mapping closure shifts to state residuals. Equation~\eqref{eq:bptt-gradient} therefore writes compactly as
\begin{equation}
\frac{\dd \mathcal{L}(\theta)}{\dd \theta}
=\frac{2}{N}\,
\underbrace{\left(\frac{\partial G_\theta}{\partial \theta}\right)^{\!\top}}_{\substack{\text{network Jacobian}\\\text{(Stage I subspace;}\\\text{Stage II backprop)}}}\;
\underbrace{\mathcal{J}_a^\top r(\theta)}_{\substack{\text{solver response}\\\text{(Stage II: }\mathcal{R}\approx\mathcal{J}_a\text{)}}}.
\label{eq:closed-loop-chain}
\end{equation}
The network Jacobian $\frac{\partial G_\theta}{\partial \theta}$ is available via automatic differentiation, whereas $\mathcal{J}_a$ is hidden inside the forward-only simulator. Because a closure output $a_k$ at step $k$ affects only future states $q_t$ ($t > k$) and never earlier steps, $\mathcal{J}_a$ has \textbf{causal block lower-triangular temporal structure} ($\mathcal{J}_{a,tk} = 0$ for $k \ge t$) governed by step-to-step error propagation. Standard black-box methods ignore both structures: ES and EKI collapse the $T$-step rollout into a static loss on $\theta \in \R^p$ and waste most perturbation energy in the null space of $\frac{\partial G_\theta}{\partial \theta}$, whereas the Neural Emulator Method \citep{frezat2026gradient} approximates $\mathcal{J}_a$ by unrolling a learned solver emulator whose state and tangent errors compound across $T$ steps.

Factorization \eqref{eq:closed-loop-chain} motivates a unified two-stage round starting from current weights $\theta_{\mathrm{c}} = (h_{\mathrm{c}}, b_{\mathrm{c}})$: \textbf{Stage~I} uses $\frac{\partial G_{\theta_{\mathrm{c}}}}{\partial \theta}$ to probe $M \ll p$ output-orthonormal weight directions and warm up the trajectory into a locally linear regime $\theta^{\mathrm{I}}$; \textbf{Stage~II} perturbs closure outputs around $\theta^{\mathrm{I}}$ to identify a structured causal linear operator $\mathcal{R} \approx \mathcal{J}_a$, solves for regularized target outputs, and applies $\big(\frac{\partial G_\theta}{\partial \theta}\big)^{\!\top}$ via network backpropagation to update all $p$ weights.

\subsection{Stage I: Parameter-Space Perturbation}
\label{sec:method-stage1}

\begin{table*}[t]
\centering
\caption{\textbf{Closed-loop test trajectory MSE on four 2D PDE benchmarks.} Mean $\pm$ s.d.\ across independent test trajectories, with changes from Coarse Physics in parentheses (\textbf{bold}/shaded: best forward-only; \underline{underlined}: runner-up). All methods start from the same initialization within each benchmark (offline $\theta_0$ on Cahn--Hilliard; $h_0 = 0$ on the other three). The bottom row reports coarse-MSE reduction recovered relative to oracle Solver BPTT.}
\label{tab:main-2d}
\fontsize{8}{9.5}\selectfont
\renewcommand{\arraystretch}{1.10}
\setlength{\tabcolsep}{3.8pt}
\begin{tabularx}{\textwidth}{@{}>{\raggedright\arraybackslash}Xcccc@{}}
\toprule
Method & \textbf{2D Cahn--Hilliard} & \textbf{2D Gray--Scott} & \textbf{2D Kolmogorov NS} & \textbf{2D Guide-Field MHD} \\
 & ($384^2\!\to\!96^2$, $\times 10^{-5}$) & ($768^2\!\to\!96^2$, $\times 10^{-3}$) & ($256^2\!\to\!64^2$, $\times 10^{-3}$) & ($256^2\!\to\!64^2$, $\times 10^{-2}$) \\
\midrule
Coarse Physics ($G_\theta \equiv 0$) & $45.2$\sd{4.4} & $125.8$\sd{4.7} & $17.07$\sd{2.09} & $6.81$\sd{1.69} \\
Frozen Offline Initialization ($G_{\theta_0}$) & $26.6$\sd{3.8} $(-41.2\%)$ & --- & --- & --- \\
\midrule
Global PPO \citep{schulman2017proximal} & $20.5$\sd{3.2} $(-54.7\%)$ & $125.6$\sd{4.7} $(-0.1\%)$ & $17.07$\sd{2.09} $(-0.0\%)$ & $6.81$\sd{1.69} $(-0.0\%)$ \\
Local PPO \citep{schulman2017proximal} & $16.1$\sd{2.7} $(-64.4\%)$ & $105.1$\sd{4.1} $(-16.5\%)$ & $17.05$\sd{2.08} $(-0.1\%)$ & $6.81$\sd{1.69} $(-0.0\%)$ \\
Closure-RL \citep{bassewitz2025closure} & $17.9$\sd{2.6} $(-60.3\%)$ & $91.3$\sd{3.5} $(-27.4\%)$ & $13.97$\sd{1.77} $(-18.2\%)$ & $6.39$\sd{1.61} $(-6.1\%)$ \\
Neural Emulator Method \citep{frezat2026gradient} & $20.6$\sd{3.3} $(-54.5\%)$ & $77.2$\sd{3.3} $(-38.6\%)$ & $6.99$\sd{0.71} $(-59.0\%)$ & $6.80$\sd{1.69} $(-0.1\%)$ \\
ES \citep{salimans2017evolution} & $11.8$\sd{2.1} $(-73.8\%)$ & $105.1$\sd{4.1} $(-16.4\%)$ & $9.97$\sd{1.11} $(-41.6\%)$ & $5.50$\sd{1.43} $(-19.2\%)$ \\
EKI \citep{kovachki2019ensemble,lopezgomez2022training} & \second{$10.3$\sd{1.8}} $(-77.3\%)$ & \second{$16.7$\sd{0.4}} $(-86.7\%)$ & \second{$6.80$\sd{0.70}} $(-60.2\%)$ & \second{$4.01$\sd{1.13}} $(-41.0\%)$ \\
\midrule
\textbf{\ourmethod{} (Ours)} & \best{$4.1$\sd{0.7} $(-90.9\%)$} & \best{$5.2$\sd{0.4} $(-95.9\%)$} & \best{$4.35$\sd{0.42} $(-74.5\%)$} & \best{$3.38$\sd{1.01} $(-50.4\%)$} \\
\midrule
\emph{Oracle Recovery (vs.\ Solver BPTT)} & \emph{$97.1\%$} & \emph{$96.6\%$} & \emph{$90.2\%$} & \emph{$87.7\%$} \\
\bottomrule
\end{tabularx}
\end{table*}

Since Stage~II relies on locally linear trajectory sensitivity around the current rollout, Stage~I first warms up the current weights $\theta_{\mathrm{c}} = (h_{\mathrm{c}}, b_{\mathrm{c}})$ into $\theta^{\mathrm{I}}$ along $M \ll p$ parameter directions: it selects $M$ directions with orthonormal output changes, adjusts the linear readout to keep finite steps aligned with those directions, and runs $2M$ perturbed rollouts to solve an $M \times M$ damped Gauss--Newton system.

\noindent\textbf{1) Select $M$ output-orthonormal parameter directions.} In a deep neural network ($p \gg M$), random weight perturbations lie mostly in the null space of $\frac{\partial G_{\theta_{\mathrm{c}}}}{\partial \theta}$ and barely move network outputs. Using forward-mode automatic differentiation of $G_{\theta_{\mathrm{c}}}$ on cached inputs $x_t$ (requiring zero solver calls), we instead construct $M$ joint weight directions $u_m = (\Delta h_m, \Delta b_m)$, $m = 1, \dots, M$, whose first-order output changes $v_m = \frac{\partial G_{\theta_{\mathrm{c}}}}{\partial \theta} u_m$ are orthonormal across the rollout.

\noindent\textbf{2) Correct the linear readout for finite-step residual.} Step~1 relies on the first-order approximation $G_{\theta_{\mathrm{c}} + \Delta\theta} \approx G_{\theta_{\mathrm{c}}} + \delta G_{\mathrm{lin}}(\Delta\theta)$, where $\delta G_{\mathrm{lin}}(\Delta\theta) = \frac{\partial G_{\theta_{\mathrm{c}}}}{\partial \theta}\Delta\theta$. Because the backbone $\Phi(b)$ is nonlinear, evaluating the network at a finite step $\Delta\theta = (\Delta h, \Delta b)$ (with $b' = b_{\mathrm{c}} + \Delta b$) deviates from $G_{\theta_{\mathrm{c}}} + \delta G_{\mathrm{lin}}(\Delta\theta)$ on cached inputs by a higher-order residual
\begin{equation}
R_{\mathrm{nl}}(\Delta\theta) = \Phi(b')\,(h_{\mathrm{c}} + \Delta h) - \big(G_{\theta_{\mathrm{c}}} + \delta G_{\mathrm{lin}}(\Delta\theta)\big).
\label{eq:bilinear-drift}
\end{equation}
Since the final layer $G_\theta(x) = \Phi(b; x)\,h$ is linear in $h$, we adjust the readout weights via least squares before calling the solver,
\begin{equation}
\Delta h_{\mathrm{comp}}(\Delta\theta) = -\Phi(b')^+\,R_{\mathrm{nl}}(\Delta\theta),
\label{eq:head-comp}
\end{equation}
where $\Phi(b')^+$ is the pseudoinverse of $\Phi(b')$, keeping finite steps aligned with $\delta G_{\mathrm{lin}}(\Delta\theta)$.

\noindent\textbf{3) Solve the $M \times M$ damped Gauss--Newton system.} For each direction $u_m$ with scale $\rho_m > 0$, we run two forward rollouts at the readout-corrected weights for $\pm\rho_m u_m$ ($2M$ rollouts in total), observe perturbed state residuals $r_{(m)}^\pm$, and form central-difference sensitivity columns $J_m = (r_{(m)}^+ - r_{(m)}^-)/(2\rho_m)$ stacked into $J = [J_1, \dots, J_M] \in \R^{N \times M}$. Solving the $M \times M$ damped Gauss--Newton system for coordinates $\xi \in \R^M$,
\begin{equation}
\left(\frac{1}{N}J^\top J + \lambda W\right)\xi = -\frac{1}{N}J^\top r(\theta_{\mathrm{c}}),
\label{eq:stage1-gn}
\end{equation}
with diagonal curvature weight $W \succ 0$ and damping $\lambda > 0$, we apply \eqref{eq:head-comp} to the combined step $\sum_{m=1}^M \xi_m u_m$ to obtain $\theta^{\mathrm{I}}$ (Appendix~\ref{app:stage1-details}).

\subsection{Stage II: Causal Output-Response Identification}
\label{sec:method-stage2}

\begin{figure*}[t]
  \centering
  \figfile[width=0.95\textwidth]{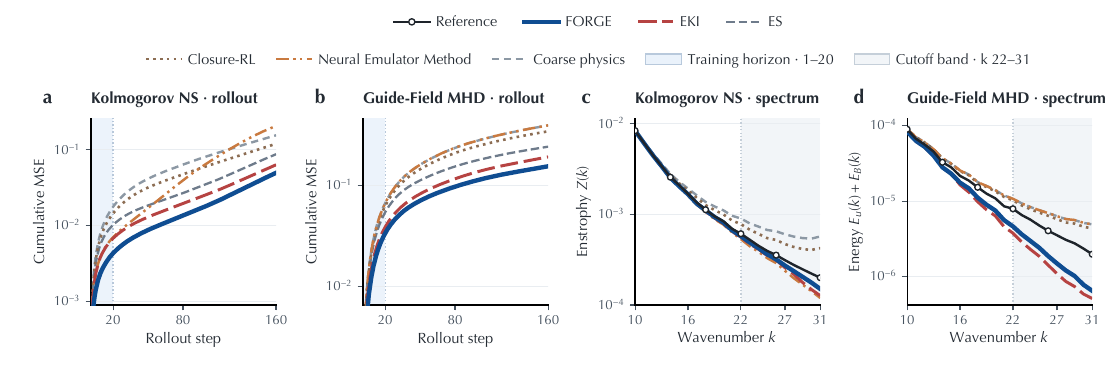}
  \caption{\textbf{Long-horizon rollout stability and high-wavenumber spectrum.} \textbf{a, b}, Cumulative validation MSE over $1$--$160$ steps ($1\times$--$8\times$ the $20$-step training window, shaded in blue) on 2D Kolmogorov NS (\textbf{a}) and guide-field MHD (\textbf{b}). \textbf{c, d}, Step-$80$ radial enstrophy spectrum $Z(k)$ (\textbf{c}) and total energy spectrum $E_u(k)+E_B(k)$ (\textbf{d}) over $k \in [10, 31]$, with gray bands marking grid-cutoff wavenumbers $k \in [22, 31]$.}
  \label{fig:long-horizon-2d}
\end{figure*}

Parameter-space probing in Stage~I updates at most $M \ll p$ directions per round, while RL and scalar zeroth-order methods collapse each trajectory into step-local or scalar rewards without cross-step state Jacobians. Stage~II instead \textbf{exploits sequential error propagation across rollout steps}: by observing how spatial output perturbations around $\theta^{\mathrm{I}}$ propagate forward in time, it identifies a causal linear response operator $\mathcal{R} \approx \mathcal{J}_a$, solves for step-wise target outputs $a_t^\star$, and updates all $p$ weights via standard backpropagation.

\noindent\textbf{1) Probe the simulator in output space.} Around the warmed-up rollout $(x_t^{\mathrm{I}}, a_t^{\mathrm{I}}, q_t^{\mathrm{I}})$ of $\theta^{\mathrm{I}}$, we re-run the simulator from $q_0$ with $K$ paired, spatially smooth, temporally orthogonal output perturbations $\pm\delta a_t$ ($K \ll N$)---injected additively in closed loop or prescribed with an identified input-response correction---and record the state differences $d_t = (q_t^{+} - q_t^{-})/2$. For locally smooth rollouts, symmetric $\pm\delta a_t$ probing cancels quadratic nonlinearity to $\mathcal{O}(\|\delta a\|^3)$, isolating the first-order trajectory response across steps.

\noindent\textbf{2) Fit a structured causal response operator $\mathcal{R}$.} By causality, $\delta a_k$ affects only future states $t > k$, yielding a block lower-triangular response $d \approx \mathcal{R}\,\delta a$:
\begin{equation}
d_t \approx (\mathcal{R}\,\delta a)_t = \sum_{k < t} \mathcal{R}_{tk}\,\delta a_k.
\label{eq:response-causal}
\end{equation}
With $K \ll N$ probes, fitting a dense $\mathcal{R}$ is underdetermined; we parameterize $\mathcal{R}$ as $d_{t+1} = A_t d_t + B_t\,\delta a_t$ using grid-matched spectral or masked differential operators augmented by state-dependent spatial modulation---modulating local pointwise and derivative couplings with the background fields $(x_t^{\mathrm{I}}, a_t^{\mathrm{I}})$ to capture the spatially varying linearized Jacobian of nonlinear advection, reactions, and interfaces---under bounded step gain $\|\mathcal{R}\,\delta a\|_2 \le \Gamma_{\max}\|\delta a\|_2$.

\noindent\textbf{3) Solve for target outputs and update all parameters.} Once $\mathcal{R}$ is fitted, updating $\theta = (h, b)$ requires zero solver calls. Approximating the perturbed residual by $r(\theta^{\mathrm{I}}) + \mathcal{R}\,\Delta a$ and minimizing $\frac{1}{N}\|r(\theta^{\mathrm{I}}) + \mathcal{R}\,\Delta a\|_2^2 + \frac{\mu}{N}\|\Delta a\|_2^2$ ($\mu > 0$) gives the damped normal equation
\begin{equation}
(\mathcal{R}^\top \mathcal{R} + \mu I)\,\Delta a = -\mathcal{R}^\top r(\theta^{\mathrm{I}}),
\label{eq:adjoint-cg}
\end{equation}
solved via conjugate gradients for target outputs $a_t^\star = a_t^{\mathrm{I}} + \Delta a_t$. These targets convert the black-box trajectory problem into supervised regression on the cached inputs $x_t^{\mathrm{I}}$, updating all $p$ weights via backpropagation through $g_\theta$ alone:
\begin{equation}
\theta_{\mathrm{proj}} = \arg\min_\theta \frac{1}{T}\sum_{t=0}^{T-1} \ell_{\mathrm{proj}}\!\big(g_\theta(x_t^{\mathrm{I}}),\, a_t^\star\big) + \lambda_\theta\,\|\theta - \theta^{\mathrm{I}}\|_2^2,
\label{eq:projection}
\end{equation}
with loss $\ell_{\mathrm{proj}}$, trust weight $\lambda_\theta > 0$, and line search along $\theta^{\mathrm{I}} + \alpha(\theta_{\mathrm{proj}} - \theta^{\mathrm{I}})$ to obtain $\theta_{\mathrm{next}}$ (Appendix~\ref{app:stage2-details}).

\begin{table*}[t]
\centering
\caption{\textbf{Closed-loop 3D POM test performance and multi-horizon forecast skill.} Four-hour test MSE, out-of-window 24-hour and 48-hour composite normalized root-mean-square error (NRMSE; $6\times$ and $12\times$ the training window), and 48-hour surface kinetic energy (KE) relative spectral error across $48$ test trajectories (mean $\pm$ s.d., with changes from Coarse Physics in parentheses).}
\label{tab:main-3d}
\fontsize{8}{9.5}\selectfont
\renewcommand{\arraystretch}{1.10}
\setlength{\tabcolsep}{5.5pt}
\begin{tabularx}{\textwidth}{@{}>{\raggedright\arraybackslash}Xcccc@{}}
\toprule
Method & \textbf{4h Test MSE} & \textbf{24h NRMSE} ($\times 10^{-2}$) & \textbf{48h NRMSE} ($\times 10^{-2}$) & \textbf{48h KE Spectral Error} \\
\midrule
Coarse Physics ($G_\theta \equiv 0$) & $1.159$\sd{0.102} & $6.02$\sd{0.36} & $8.67$\sd{0.56} & $9.58\%$ \\
\midrule
EKI & $1.159$\sd{0.102} $(-0.0\%)$ & $6.02$\sd{0.36} $(+0.0\%)$ & $8.67$\sd{0.56} $(+0.0\%)$ & $9.58\%$ $(+0.0\%)$ \\
PPO & $1.158$\sd{0.102} $(-0.1\%)$ & $6.02$\sd{0.36} $(-0.0\%)$ & $8.67$\sd{0.56} $(-0.0\%)$ & $9.58\%$ $(-0.0\%)$ \\
ES & $1.132$\sd{0.102} $(-2.3\%)$ & $5.94$\sd{0.37} $(-1.4\%)$ & $8.58$\sd{0.57} $(-1.1\%)$ & \second{$9.57\%$} $(-0.1\%)$ \\
Closure-RL & $1.118$\sd{0.101} $(-3.5\%)$ & $5.94$\sd{0.37} $(-1.4\%)$ & $8.54$\sd{0.56} $(-1.5\%)$ & $9.82\%$ $(+2.4\%)$ \\
Neural Emulator Method & \second{$0.614$\sd{0.051}} $(-47.0\%)$ & \second{$5.12$\sd{0.35}} $(-14.9\%)$ & \second{$7.72$\sd{0.53}} $(-10.9\%)$ & $9.77\%$ $(+1.9\%)$ \\
\midrule
\textbf{\ourmethod{} (Ours)} & \best{$0.347$\sd{0.029} $(-70.1\%)$} & \best{$3.88$\sd{0.27} $(-35.5\%)$} & \best{$7.39$\sd{0.46} $(-14.8\%)$} & \best{$8.41\%$ $(-12.3\%)$} \\
\midrule
\emph{Solver BPTT (oracle)} & \emph{$0.332$\sd{0.029} $(-71.3\%)$} & \emph{$5.47$\sd{0.19} $(-9.1\%)$} & \emph{$9.27$\sd{0.28} $(+6.8\%)$} & \emph{$2.77\%$ $(-71.1\%)$} \\
\bottomrule
\end{tabularx}
\end{table*}

\begin{figure*}[t]
  \centering
  \figfile[width=0.95\textwidth]{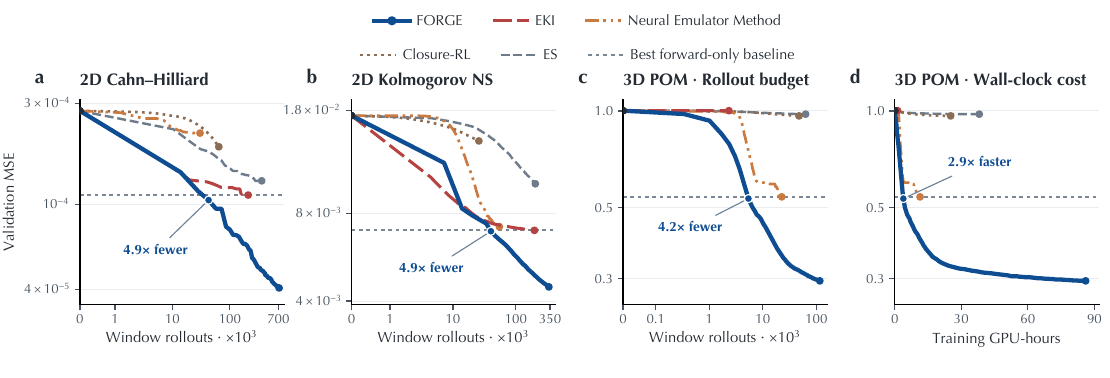}
  \caption{\textbf{Query efficiency and matched-hardware wall-clock cost.} \textbf{a--c}, Validation MSE vs.\ cumulative training window rollouts from shared initialization on 2D Cahn--Hilliard (\textbf{a}), Kolmogorov NS (\textbf{b}), and 3D POM (\textbf{c}); dots mark stopping budgets. \textbf{d}, Validation MSE vs.\ matched-hardware GPU-hours on 3D POM \citep{shu2026neuralpom}. Dashed lines mark the best forward-only baseline; blue circles mark \ourmethod{} crossovers.}
  \label{fig:efficiency}
\end{figure*}

\subsection{Theoretical Analysis}
\label{sec:method-theory}

\ourmethod{} replaces unavailable solver derivatives in \eqref{eq:closed-loop-chain} with readout-corrected parameter probing in Stage~I and causal response inversion in Stage~II. Propositions~\ref{prop:stage1-cancel} and~\ref{prop:stage2-descent} show how these two designs ensure accurate subspace warm-up and stable full-network descent from forward runs alone (proofs in Appendix~\ref{app:proofs}).

In Stage~I, finite-step backbone nonlinearity would otherwise corrupt central-difference sensitivities $J_m$. Proposition~\ref{prop:stage1-cancel} shows that readout correction \eqref{eq:head-comp} eliminates higher-order residuals within the readout's feature span while preserving the exact first-order network Jacobian, enabling accurate Gauss--Newton warm-up steps.

\begin{proposition}[Nonlinear Residual Cancellation in Stage~I]
\label{prop:stage1-cancel}
Suppose $\Phi$ is $C^2$ near $b_{\mathrm{c}}$ and $\Phi(b')^+$ is locally bounded near $b_{\mathrm{c}}$ with orthogonal projector $\Pi = \Phi(b')\Phi(b')^+$. For a finite step $\Delta\theta = (\Delta h, \Delta b)$ to $b' = b_{\mathrm{c}} + \Delta b$, let $\Delta G(\Delta\theta) = \Phi(b')(h_{\mathrm{c}} + \Delta h + \Delta h_{\mathrm{comp}}(\Delta\theta)) - G_{\theta_{\mathrm{c}}}$ under \eqref{eq:head-comp}. Then $\left.\frac{\partial (\Delta G)}{\partial \Delta\theta}\right|_{\Delta\theta=0} = \frac{\partial G_{\theta_{\mathrm{c}}}}{\partial \theta}$, and
\begin{equation}
\Delta G(\Delta\theta) - \delta G_{\mathrm{lin}}(\Delta\theta) = (I - \Pi)\,R_{\mathrm{nl}}(\Delta\theta).
\label{eq:drift-cancellation}
\end{equation}
\end{proposition}
Since $\Pi(I - \Pi) = 0$, the nonlinear residual in $\mathrm{range}(\Pi)$ vanishes, keeping finite-step probes faithful to the output-orthonormal tangents and yielding reliable loss reduction in Stage~I.

In Stage~II, $\mathcal{R}$ is fitted from only $K \ll N$ rollouts and is thus an inexact estimate of $\mathcal{J}_a$. Proposition~\ref{prop:stage2-descent} shows why this approximate operator still drives effective full-network optimization: damped inversion \eqref{eq:adjoint-cg} keeps target updates bounded regardless of small singular values in $\mathcal{R}$, and a finite accuracy margin suffices to guarantee closed-loop descent without recovering $\mathcal{J}_a$ exactly.

\begin{proposition}[Target Stability and Closed-Loop Descent in Stage~II]
\label{prop:stage2-descent}
Let $r_{\mathrm{I}} = r(\theta^{\mathrm{I}})$, let $\mathcal{J}_a$ be the closed-loop trajectory Jacobian at $a^{\mathrm{I}}$, and let $\Delta a$ solve \eqref{eq:adjoint-cg} with damping $\mu > 0$ for a bounded operator $\mathcal{R}$ ($\|\mathcal{R}\|_2 \le \Gamma_{\max} < \infty$).
\textbf{(i)~Bounded target update:} Independent of $\Gamma_{\max}$, $\Delta a$ and its predicted response satisfy
\begin{equation}
\|\Delta a\|_2 \le \frac{1}{2\sqrt{\mu}}\,\|r_{\mathrm{I}}\|_2, \qquad \|\mathcal{R}\,\Delta a\|_2 \le \|r_{\mathrm{I}}\|_2.
\label{eq:response-gain}
\end{equation}
\textbf{(ii)~First-order closed-loop descent:} If $\mathcal{R}$ approximates $\mathcal{J}_a$ along $\Delta a$ with relative error $\|(\mathcal{J}_a-\mathcal{R})\Delta a\|_2\le\epsilon\,\|\mathcal{J}_a\Delta a\|_2$ for $\epsilon \in (0, 1)$, then $\langle\mathcal{J}_a\Delta a, r_{\mathrm{I}}\rangle < 0$ whenever
\begin{equation}
\|\mathcal{R}\,\Delta a\|_2 > \frac{\epsilon}{1-\epsilon}\,\|r_{\mathrm{I}}\|_2.
\label{eq:adjoint-descent}
\end{equation}
\end{proposition}
Intuitively, \eqref{eq:adjoint-descent} shows that whenever the predicted residual correction $\|\mathcal{R}\,\Delta a\|_2$ exceeds the approximation-error floor $\frac{\epsilon}{1-\epsilon}\|r_{\mathrm{I}}\|_2$, $\Delta a$ stays a genuine closed-loop descent direction ($\langle\mathcal{J}_a\Delta a, r_{\mathrm{I}}\rangle < 0$) for updating all $p$ weights in \eqref{eq:projection}.

\section{Experiments}
\label{sec:experiments}

\subsection{Experimental Setup}
\label{sec:exp-setup}

We evaluate \ourmethod{} on five hybrid PDE benchmarks (Fig.~\ref{fig:benchmarks}; Appendix~\ref{app:benchmarks}). In each benchmark, reference trajectories are simulated on a fine grid, downsampled onto a coarse grid, and split into disjoint training, validation, and test trajectories before extracting multi-step rollout windows. Training and evaluation run the coarse solver, so the embedded network must learn the subgrid closure bridging coarse and fine dynamics. Four periodic 2D benchmarks pair coarse PDE solvers with Fourier neural operator (FNO) closures for phase separation (Cahn--Hilliard), reaction--diffusion fronts (Gray--Scott), forced fluid turbulence (Kolmogorov NS), and magnetohydrodynamic turbulence (guide-field MHD). A fifth bounded 3D benchmark embeds a convolutional closure predicting baroclinic momentum and tracer tendencies inside a semi-implicit free-surface ocean circulation solver \citep{blumberg1987description,shu2026neuralpom}.

Throughout training of \ourmethod{} and all forward-only baselines, each simulator is treated as a forward-only black box without internal solver knowledge: the four 2D solvers run in external CPU processes, and 3D NeuralPOM \citep{shu2026neuralpom} runs on GPU with autograd disabled so only forward states are returned. Within each benchmark, \ourmethod{} and all baselines start from the same initialization $\theta_0$. We compare against \textbf{ES} \citep{salimans2017evolution}, \textbf{EKI} \citep{kovachki2019ensemble,lopezgomez2022training}, \textbf{Global/Local PPO} \citep{schulman2017proximal,novati2021automating}, \textbf{Closure-RL} \citep{bassewitz2025closure}, and the \textbf{Neural Emulator Method} \citep{frezat2026gradient} (which differentiates through a learned step emulator to train $G_\theta$; baseline protocols in Appendix~\ref{app:baselines}), alongside \textbf{Solver BPTT} \eqref{eq:bptt-gradient} as a white-box oracle. Online Euler gradient approximations \citep{ouala2024online} target direct additive state corrections or require tangent-linear/state-ensemble propagation, and thus cannot be applied directly to our embedded forward-only closures (Section~\ref{sec:exp-convergence}; Appendix~\ref{app:ega}).

\begin{table*}[t]
\centering
\caption{\textbf{Ablation of Stage~I and Stage~II components across all five benchmarks.} Closed-loop test MSE (mean $\pm$ s.d.) under the protocol of Tables~\ref{tab:main-2d}--\ref{tab:main-3d}; shaded row marks full \ourmethod{} (\textbf{bold}: best), followed by stage ($1$--$2$) and mechanism ($3$--$6$) ablations.}
\label{tab:ablation}
\fontsize{8}{9.5}\selectfont
\renewcommand{\arraystretch}{1.12}
\setlength{\tabcolsep}{3.5pt}
\begin{tabularx}{\textwidth}{@{}>{\raggedright\arraybackslash}Xccccc@{}}
\toprule
 & \textbf{2D Cahn--} & \textbf{2D Gray--} & \textbf{2D Kolmogorov} & \textbf{2D Guide-Field} & \textbf{3D POM} \\
Variant & \textbf{Hilliard} ($\times\!10^{-5}$) & \textbf{Scott} ($\times\!10^{-3}$) & \textbf{NS} ($\times\!10^{-3}$) & \textbf{MHD} ($\times\!10^{-2}$) & (4h Test) \\
\midrule
\textbf{\ourmethod{} (Ours, Stage~I + Stage~II)} & \best{$4.13$\sd{0.70}} & \best{$5.22$\sd{0.37}} & \best{$4.35$\sd{0.42}} & \best{$3.38$\sd{1.01}} & \best{$0.347$\sd{0.029}} \\
\midrule
1.\ w/o Stage~I Warm-Up (Stage-II-Only) & $9.81$\sd{1.45} & $96.37$\sd{4.19} & $17.07$\sd{2.09} & $3.48$\sd{1.05} & $0.361$\sd{0.028} \\
2.\ w/o Stage~II Output Perturbation (Stage-I-Only) & $5.79$\sd{0.92} & $9.40$\sd{0.22} & $6.40$\sd{0.63} & $3.76$\sd{1.07} & $1.153$\sd{0.102} \\
\midrule
3.\ Stage~I w/o Readout Compensation ($\Delta h_{\mathrm{comp}}\!=\!0$) & $4.35$\sd{0.69} & $15.05$\sd{0.43} & $5.46$\sd{0.53} & $3.43$\sd{0.99} & $0.366$\sd{0.028} \\
4.\ Stage~II w/o State-Dependent Coupling & $5.79$\sd{0.92} & $5.36$\sd{0.36} & $6.51$\sd{0.66} & $3.77$\sd{1.06} & $0.367$\sd{0.028} \\
5.\ Stage~II w/o Multi-Step Calibration & $4.36$\sd{0.70} & $5.46$\sd{0.37} & $5.04$\sd{0.52} & $3.59$\sd{0.98} & $0.356$\sd{0.029} \\
6.\ Stage~II Head-Only Update (Freeze $b$) & $5.77$\sd{0.93} & $5.68$\sd{0.40} & $6.27$\sd{0.61} & $4.65$\sd{1.21} & $0.727$\sd{0.071} \\
\bottomrule
\end{tabularx}
\end{table*}

\subsection{Closed-Loop Accuracy}
\label{sec:exp-accuracy}

\ourmethod{} achieves the lowest test MSE among all forward-only methods on all four 2D PDEs (Table~\ref{tab:main-2d}) and on 3D POM (Table~\ref{tab:main-3d}), reducing the rollout error of Coarse Physics by $50\%$--$96\%$ in 2D and $70\%$ in 3D. In 2D, \ourmethod{} lowers test MSE by $16\%$--$69\%$ below EKI, the strongest forward-only baseline across all four regimes. On 3D POM, weight-space zeroth-order and reinforcement-learning methods stay within $4\%$ of Coarse Physics, whereas \ourmethod{} lowers test MSE by $44\%$ below the Neural Emulator Method.

Without solver derivatives, \ourmethod{} recovers $88\%$--$98\%$ of the coarse-solver MSE reduction achieved by converged Solver BPTT across the five benchmarks (Tables~\ref{tab:main-2d}--\ref{tab:main-3d}). On 4-hour 3D POM, \ourmethod{} comes within $5\%$ of Solver BPTT in total MSE and attains lower error on velocity and temperature (Appendix~\ref{app:visualizations}).

\subsection{Long-Horizon Stability and Spectral Behavior}
\label{sec:exp-physics}

Closures are trained on short windows ($20$ steps in 2D; $4$ hours in 3D); for long-term validation, we roll each learned closure out to $80$ and $160$ steps in 2D and $48$ hours in 3D. Across Kolmogorov NS, guide-field MHD, and Cahn--Hilliard, all \ourmethod{} rollouts remain stable through $160$ steps, lowering cumulative long-horizon MSE by $55\%$--$78\%$ below Coarse Physics and $17\%$--$62\%$ below EKI (Fig.~\ref{fig:long-horizon-2d}a,b). Radial spectra reveal why (Fig.~\ref{fig:long-horizon-2d}c,d). On a coarse grid, unresolved small scales fail to dissipate high-frequency modes, causing energy to pile up near the grid cutoff ($k \in [22, 31]$). RL and the Neural Emulator Method do not consistently remove this buildup, whereas \ourmethod{} suppresses cutoff accumulation, closely tracking the reference spectrum on Kolmogorov NS and eliminating the high-wavenumber pile-up on MHD.

This long-term stability extends to realistic 3D baroclinic ocean dynamics in POM, coupling free-surface waves, stratification, and 3D currents across $15$ layers (Table~\ref{tab:main-3d}). Over 48-hour forecasts, Solver BPTT fits the 4-hour training window tightly and reaches the lowest surface kinetic energy spectral error ($2.77\%$), yet slow sea-surface drift raises its 48-hour composite NRMSE above Coarse Physics. By contrast, \ourmethod{} remains stable through 48 hours, reducing 24-hour and 48-hour composite NRMSE by $35.5\%$ and $14.8\%$ and 48-hour spectral error by $12.3\%$.

\subsection{Query Efficiency and Response Accuracy}
\label{sec:exp-convergence}

Against cumulative solver rollouts from the shared initialization (Fig.~\ref{fig:efficiency}a--c), \ourmethod{} reaches the final validation error of the strongest forward-only baseline using $42\%$--$80\%$ fewer rollouts across the five benchmarks, and achieves lower error at every baseline's stopping budget. In matched-hardware wall-clock time on 3D POM (Fig.~\ref{fig:efficiency}d), \ourmethod{} reaches the Neural Emulator Method's final accuracy in one third the GPU-hours ($3.9$ vs.\ $11.4\text{ GPU-h}$) and continues down to $0.296$ validation MSE ($86.0\text{ GPU-h}$ total).

To see why Stage~II guides updates more accurately than the Neural Emulator Method, we apply unseen output perturbations $\delta a$ to the simulator and compare each method's predicted trajectory shift against the true solver difference (Appendix~\ref{app:closed-loop-check}). Because the Neural Emulator Method unrolls a learned step model autoregressively over $100$--$480$ substeps per 2D window, small single-step errors snowball into large trajectory and gradient drift; \ourmethod{} instead fits its structured operator $\mathcal{R}$ directly on window-level perturbations around the current rollout, matching true solver responses with $0.89$--$0.98$ median cosine similarity. Meanwhile, online Euler gradient approximations \citep{ouala2024online} do not directly apply here: their static form assumes additive state tendencies ($\partial q_{t+1}/\partial a_t \approx \Delta t\,I$) that omit spatial or partial-state operators inside $H$, while their tangent-linear and ensemble variants rely on solver derivatives or unstructured state-space propagation (Appendix~\ref{app:ega}).

\subsection{Ablation and Mechanism Analysis}
\label{sec:exp-ablation}

\textbf{Complementary roles of Stage~I and Stage~II.} Both stages are needed for reliable optimization (Table~\ref{tab:ablation}, rows~1--2). Stage~II fits a local linear operator $\mathcal{R}$ around the current rollout, requiring trajectories where finite output perturbations behave nearly linearly. Without Stage~I warm-up (row~1), starting directly from $\theta_0$ leaves strongly nonlinear trajectories far from the reference, causing linear response fits in Stage~II to stall or degrade on Gray--Scott, Kolmogorov NS, and Cahn--Hilliard; even where initial rollouts behave more smoothly (MHD and 4-hour 3D POM), Stage~I warm-up lowers test error. Conversely, running Stage~I alone (row~2) restricts each round to $M \ll p$ parameter directions, underfitting the 2D closures and barely moving the 3D ocean network ($1.153$ vs.\ $0.347$).

\textbf{Drift compensation and response operator structure.} Dropping readout drift compensation in Stage~I ($\Delta h_{\mathrm{comp}} = 0$, row~3) allows nonlinear hidden-layer feature shifts to corrupt finite-difference sensitivities (Proposition~\ref{prop:stage1-cancel}), raising error across all five benchmarks and nearly tripling error on Gray--Scott. Within Stage~II, removing state-dependent modulation (row~4) leaves a spatially homogeneous operator that cannot track local interface or advective flow variations, stalling near or slightly above Stage-I-Only on Cahn--Hilliard, Kolmogorov NS, and MHD while degrading Gray--Scott and 3D POM. Fitting $\mathcal{R}$ on single-step differences alone (row~5) improves over Stage-I-Only across all five systems yet consistently trails full multi-step calibration, confirming that matching cross-step error propagation over the rollout window sharpens target recovery.

\textbf{Full-network backpropagation.} Updating only the linear readout $h$ while freezing the hidden backbone $b$ in Stage~II (row~6) substantially degrades accuracy across all five benchmarks, yielding minimal gain over Stage-I-Only on Cahn--Hilliard and Kolmogorov NS, worsening error on coupled MHD, and leaving nearly half the reducible error on 3D POM ($0.727$ vs.\ $0.347$). Because $h$ accounts for under $0.4\%$ of network weights and is already adjusted in Stage~I, refitting $h$ alone on fixed features $\Phi(b^{\mathrm{I}})$ cannot adapt hidden representations, whereas backpropagating through $b$ updates all layers without extra solver calls.

\textbf{Closed-loop tangent alignment.} Across twenty rounds on Kolmogorov NS, the trajectory shift $\mathcal{R}\,\delta a(\delta\theta)$ induced by Stage~II's full-network update $\delta\theta = \theta_{\mathrm{proj}} - \theta^{\mathrm{I}}$ matches the exact Solver BPTT tangent $\mathcal{J}_a\,\delta a(\delta\theta)$ with $0.91$ median cosine similarity (Appendix~\ref{app:closed-loop-check}).

\section{Conclusion}
\label{sec:conclusion}

We introduced a five-benchmark forward-only hybrid PDE suite and \ourmethod{}, an online framework for training embedded neural closures without solver derivatives. Decoupling exact neural network Jacobians from causal multi-step simulator responses, \ourmethod{} warms up rollouts in an output-orthonormal weight subspace with readout drift compensation (Stage~I) and updates all weights by fitting a structured error-propagation operator in spatial output space followed by network backpropagation (Stage~II). Across four 2D PDEs and a 3D primitive-equation ocean circulation solver \citep{shu2026neuralpom} under forward-only execution, \ourmethod{} outperforms zeroth-order, ensemble, reinforcement-learning, and emulator-based baselines, matches the best forward-only baseline using $42\%$--$80\%$ fewer rollouts, recovers $88\%$--$98\%$ of Solver BPTT's gains, and stays stable up to $8\times$--$12\times$ the training horizon.

Looking ahead, this forward-only approach enables operational Earth-system, engineering, and multiphysics simulators to learn embedded parameterizations directly from multi-step trajectories without adjoint development. Future work will extend causal response identification to coupled climate models, unstructured meshes, and sparse observations.

\subsubsection*{Acknowledgments}
This work was supported by the Fundamental and Interdisciplinary Disciplines Breakthrough Plan of the Ministry of Education of China (JYB2025XDXIM801) and the National Natural Science Foundation of China (42125503 and 42430602).

{\small
\setlength{\bibsep}{3.0pt plus 0.8pt minus 0.5pt}
\putbib[refs]
}
\end{bibunit}

\clearpage
\appendix
\numberwithin{equation}{section}
\numberwithin{table}{section}
\numberwithin{figure}{section}
\begin{bibunit}[plainnat]
\makeatletter
\def\hyper@natlinkstart#1{\hyper@linkstart{cite}{appcite.#1}\def\hyper@nat@current{#1}}
\def\hyper@natlinkbreak#1#2{\hyper@linkend#1\hyper@linkstart{cite}{appcite.#2}}
\def\hyper@natanchorstart#1{\Hy@raisedlink{\hyper@anchorstart{appcite.#1}}}
\makeatother
\makeatletter
\setlength{\@fptop}{0pt}\setlength{\@dblfptop}{0pt}
\setlength{\@fpsep}{14pt}\setlength{\@dblfpsep}{12pt}
\setlength{\@fpbot}{0pt plus 1fil}\setlength{\@dblfpbot}{0pt plus 1fil}
\makeatother
\setlength{\dblfloatsep}{10pt plus 2pt minus 2pt}
\setlength{\dbltextfloatsep}{10pt plus 2pt minus 2pt}
\raggedbottom

\section*{Overview}
This supplementary material is organized into four sections.
\textbf{Appendix~\ref{app:method-details}} collects our notation, outlines one optimization round of \ourmethod{}, proves the theoretical results in Section~\ref{sec:method}, and details the Stage~I and Stage~II implementations.
\textbf{Appendix~\ref{app:benchmarks}} describes the five hybrid PDE benchmarks, governing equations, and evaluation metrics.
\textbf{Appendix~\ref{app:baselines}} covers baseline setups, rollout accounting, and Euler gradient approximations.
\textbf{Appendix~\ref{app:extended-results}} reports oracle verification, cold-start and long-horizon results, rollout visualizations, and response diagnostics.

\section{Proofs and Algorithmic Details}
\label{app:method-details}

Table~\ref{tab:app-notation} summarizes our notation, and Algorithm~\ref{alg:forge} outlines one optimization round of \ourmethod{}.

\begin{table*}[!htb]
\centering
\caption{\textbf{Summary of notation used in the derivations and algorithm.}}
\label{tab:app-notation}
\footnotesize
\renewcommand{\arraystretch}{1.06}
\setlength{\tabcolsep}{5pt}
\begin{tabularx}{\textwidth}{@{}>{\raggedright\arraybackslash}p{0.27\textwidth}>{\raggedright\arraybackslash}X@{}}
\toprule
Symbol & Meaning \\
\midrule
$q_t$, $q_t^{\mathrm{ref}}$, $x_t$, $a_t$ & Coarse solver state, reference state, closure input $x_t = F(q_t)$, and closure output $a_t = g_\theta(x_t) = g_0(x_t) + G_\theta(x_t)$ at step $t$. \\
$F$, $g_0$, $G_\theta$, $H$ & Pre-closure solver map, baseline closure, neural closure, and post-closure time stepper. \\
$r(\theta)$, $r_{\mathrm{I}}$, $\mathcal{L}(\theta)$, $T$, $N_{\mathrm{cb}}$ & Stacked normalized state errors $q_t(\theta) - q_t^{\mathrm{ref}}$, residual $r(\theta^{\mathrm{I}})$ after Stage~I, rollout MSE $\frac{1}{N}\|r(\theta)\|_2^2$, scored steps, and total closure calls per window. \\
$\theta = (h, b)$, $\Phi(b; x)$ & Network weights split into linear readout $h$ and hidden backbone $b$ ($p$ total weights), with penultimate features $\Phi(b; x)$ so $G_\theta(x) = \Phi(b; x)\,h$. \\
$\theta_0$, $\theta_{\mathrm{c}}$, $\theta^{\mathrm{I}}$, $\theta_{\mathrm{proj}}$, $\theta_{\mathrm{next}}$ & Shared initial weights, round-start weights, Stage~I weights, projected Stage~II weights, and accepted round-end weights. \\
$u_m$, $v_m$, $\rho_m$, $M$ & Stage~I weight directions ($m = 1, \dots, M$), first-order output tangents $v_m = \frac{\partial G_{\theta_{\mathrm{c}}}}{\partial\theta}u_m$, probe radii, and subspace dimension ($M \ll p$). \\
$R_{\mathrm{nl}}$, $\Delta h_{\mathrm{comp}}$, $\Pi$ & Higher-order network remainder \eqref{eq:bilinear-drift}, readout drift correction \eqref{eq:head-comp} (rank-$s$ truncated SVD), and orthogonal projector $\Pi = \Phi(b')\Phi(b')^+$. \\
$J$, $W$, $\lambda$, $\xi$ & Stage~I central-difference sensitivity matrix, diagonal curvature weight, damping parameter, and subspace step coordinates \eqref{eq:stage1-gn}. \\
$\mathcal{J}_{\mathrm{ol}}$, $\mathcal{K}$, $\mathcal{J}_{\mathrm{cb}}$, $\mathcal{P}$, $\mathcal{J}_a$ & Open-loop closure-call Jacobian, call-to-input feedback Jacobian, closed-loop closure-call Jacobian, substep prolongation operator, and interval-level Jacobian $\mathcal{J}_a = \mathcal{J}_{\mathrm{cb}}\mathcal{P}$. \\
$\delta a$, $d$, $K$, $\mathcal{R}$, $\Gamma_{\max}$ & Stage~II output perturbation (RMS $\sigma$), state difference, probe-pair count ($K \ll N$), fitted causal response operator, and global gain bound. \\
$A_t$, $B_t$, $\delta x$ & 2D linear recurrence operators in $d_{t+1} = A_t d_t + B_t\,\delta a_t$, and predicted closure input shift in 3D POM. \\
$\Delta a$, $a_t^\star$, $\mu$, $\tau_a$ & Target output update from \eqref{eq:adjoint-cg}, step-wise target $a_t^\star = a_t^{\mathrm{I}} + \Delta a_t$, Tikhonov damping, and trust-region radius. \\
$\epsilon$, $\mathcal{M}$, $e_{\mathrm{cb}}$ & Relative operator error along $\Delta a$, descent margin \eqref{eq:app-descent-margin}, and closure-call tangent residual after weight projection. \\
\bottomrule
\end{tabularx}
\end{table*}

\begin{algorithm}[!htb]
\small
\DontPrintSemicolon
\caption{One Optimization Round of \ourmethod{}}
\label{alg:forge}
\KwIn{Current weights $\theta_{\mathrm{c}} = (h_{\mathrm{c}}, b_{\mathrm{c}})$ and training windows $\{q_0^{(w)}, q_{1:T}^{\mathrm{ref},(w)}\}$.}
\textbf{Stage I: Warm up the rollout in a small weight subspace}\;
Build $M \ll p$ joint readout--backbone directions $u_m$ spanning diverse output variations.\;
Run $2M$ forward rollouts at $\pm\rho_m u_m$ with readout compensation \eqref{eq:head-comp} to cancel nonlinear feature drift.\;
Form finite-difference sensitivities $J$, solve the $M \times M$ Gauss--Newton system \eqref{eq:stage1-gn}, and screen candidate steps on forward rollouts to obtain $\theta^{\mathrm{I}}$ (keep $\theta_{\mathrm{c}}$ if none improves the loss).\;
\textbf{Stage II: Fit a causal output response and update all weights}\;
Run the nominal rollout at $\theta^{\mathrm{I}}$ to record states, inputs, outputs, and residual $r(\theta^{\mathrm{I}})$.\;
Inject paired output perturbations $\pm\delta a_t$ from $q_0^{(w)}$, fit the causal response operator $\mathcal{R}$ (plus input-shift map $\delta x$ in 3D), and validate $\mathcal{R}$ on disjoint held-out probes.\;
Solve \eqref{eq:adjoint-cg} with trust-region CG for target outputs $a_t^\star = a_t^{\mathrm{I}} + \Delta a_t$, then update all weights $\theta$ by backpropagating through $G_\theta$ alone ($H^2$ variable projection in 2D; cached-window readout initialization and response-weighted refinement in 3D).\;
Evaluate candidate step scales on forward rollouts and keep the lowest-loss update improving on $\theta^{\mathrm{I}}$ as $\theta_{\mathrm{next}}$.\;
\KwOut{Updated weights $\theta_{\mathrm{next}}$.}
\end{algorithm}

\subsection{Theoretical Proofs}
\label{app:proofs}

\paragraph{Why the closed-loop gradient factors as \eqref{eq:closed-loop-chain}.}
Consider an infinitesimal parameter perturbation $\delta\theta$ during a rollout with $N_{\mathrm{cb}}$ closure calls (including all numerical substeps). Changing $\theta$ alters the closure outputs directly, which changes later solver states and therefore later closure inputs; those shifted inputs then feed back through the closure again.

To write this feedback loop cleanly, stack all $N_{\mathrm{cb}}$ closure inputs $x$ and outputs $a$ across the window. Let $\mathcal{J}_{\mathrm{ol}}$ map prescribed output shifts to final trajectory residual shifts $\delta r$, and let $\mathcal{K}$ map output shifts to subsequent input shifts $\delta x$. For the closure itself, $\frac{\partial G_\theta}{\partial \theta}$ is the direct derivative on fixed inputs, and $\frac{\partial g_\theta}{\partial x} = \frac{\partial g_0}{\partial x} + \frac{\partial G_\theta}{\partial x}$ is the block-diagonal input derivative. Linearizing the rollout around $\theta$ gives
\begin{equation}
\begin{aligned}
\delta a_{\mathrm{tot}} &= \frac{\partial G_\theta}{\partial \theta}\,\delta\theta + \frac{\partial g_\theta}{\partial x}\,\delta x, \qquad \delta x = \mathcal{K}\,\delta a_{\mathrm{tot}},\\
\delta r &= \mathcal{J}_{\mathrm{ol}}\,\delta a_{\mathrm{tot}}.
\end{aligned}
\label{eq:app-feedback-linearization}
\end{equation}

Because time marches forward, a closure call can affect only strictly later inputs. Hence $\mathcal{K}$ is strictly block lower-triangular, which means $\bigl(\frac{\partial g_\theta}{\partial x}\,\mathcal{K}\bigr)^{N_{\mathrm{cb}}} = 0$. The feedback matrix $I - \frac{\partial g_\theta}{\partial x}\,\mathcal{K}$ is therefore always invertible via a finite sum:
\begin{equation}
\left(I - \frac{\partial g_\theta}{\partial x}\,\mathcal{K}\right)^{-1} = \sum_{j=0}^{N_{\mathrm{cb}}-1}\left(\frac{\partial g_\theta}{\partial x}\,\mathcal{K}\right)^j.
\label{eq:app-feedback-inverse}
\end{equation}

Now consider injecting an additive perturbation $\delta a_{\mathrm{cb}}$ into the closure outputs while keeping $g_\theta$ active in closed loop. The same feedback balance reads $\delta a_{\mathrm{tot}} = \delta a_{\mathrm{cb}} + \frac{\partial g_\theta}{\partial x}\,\mathcal{K}\,\delta a_{\mathrm{tot}}$, so the closed-loop sensitivity of the residual to output perturbations is
\begin{equation}
\mathcal{J}_{\mathrm{cb}} = \mathcal{J}_{\mathrm{ol}}\left(I - \frac{\partial g_\theta}{\partial x}\,\mathcal{K}\right)^{-1}.
\label{eq:app-closed-loop-response}
\end{equation}
Substituting $\delta x = \mathcal{K}\,\delta a_{\mathrm{tot}}$ into \eqref{eq:app-feedback-linearization} and comparing with \eqref{eq:app-closed-loop-response} gives $\frac{\dd r}{\dd\theta} = \mathcal{J}_{\mathrm{cb}}\frac{\partial G_\theta}{\partial \theta}$. Differentiating $\mathcal{L}(\theta) = \frac{1}{N}\|r(\theta)\|_2^2$ immediately recovers the factored gradient \eqref{eq:closed-loop-chain}.

Stage~II perturbs outputs at the coarser interval level: one shift per scored output step across the internal substeps in 2D ($N_{\mathrm{cb}} > T$), or one per hourly closure call in 3D POM ($N_{\mathrm{cb}} = 4$, where the $1/6$ substep hold resides inside $H$ in \eqref{eq:app-pom-discrete}). Denoting this piecewise-constant prolongation across closure calls by $\mathcal{P}$ ($\mathcal{P} = I$ on the four hourly calls in 3D), the interval-level Jacobian in Section~\ref{sec:method} is $\mathcal{J}_a = \mathcal{J}_{\mathrm{cb}}\mathcal{P}$ (coinciding with $\mathcal{J}_{\mathrm{cb}}$ whenever $N_{\mathrm{cb}} = T$). In 2D, probing with $g_{\theta^{\mathrm{I}}}$ active identifies $\mathcal{J}_a$ directly; in 3D, open-loop hourly probes identify $\mathcal{J}_{\mathrm{ol}}$ while a companion linear model estimates $\mathcal{K}$ so input feedback propagates through $G_\theta$.

Finally, this factorization explains why isotropic random weight perturbations stall in deep closures: if $\frac{\partial G_\theta}{\partial \theta}$ has rank $d_G \ll p$, a random unit vector in $\R^p$ places $(p - d_G)/p$ of its expected squared norm in the null space of $\frac{\partial G_\theta}{\partial \theta}$ and barely changes the closure output.

\begin{proof}[Proof of Proposition~\ref{prop:stage1-cancel}]
Consider a finite weight step $\Delta\theta = (\Delta h, \Delta b)$ around $\theta_{\mathrm{c}} = (h_{\mathrm{c}}, b_{\mathrm{c}})$ on the cached inputs, with $b' = b_{\mathrm{c}} + \Delta b$. Expanding the compensated output change $\Delta G(\Delta\theta) = \Phi(b')(h_{\mathrm{c}} + \Delta h + \Delta h_{\mathrm{comp}}(\Delta\theta)) - G_{\theta_{\mathrm{c}}}$ and substituting the remainder \eqref{eq:bilinear-drift} and readout correction $\Delta h_{\mathrm{comp}}(\Delta\theta) = -\Phi(b')^+ R_{\mathrm{nl}}(\Delta\theta)$ from \eqref{eq:head-comp} gives
\begin{equation*}
\begin{aligned}
\Delta G(\Delta\theta) - \delta G_{\mathrm{lin}}(\Delta\theta) &= R_{\mathrm{nl}}(\Delta\theta) + \Phi(b')\Delta h_{\mathrm{comp}}(\Delta\theta) \\
&= (I - \Pi)\,R_{\mathrm{nl}}(\Delta\theta),
\end{aligned}
\end{equation*}
where $\Pi = \Phi(b')\Phi(b')^+$ is the orthogonal projector onto the retained feature subspace. This establishes \eqref{eq:drift-cancellation} and shows that the component of $R_{\mathrm{nl}}(\Delta\theta)$ in $\mathrm{range}(\Pi)$ vanishes identically.

Because $\Phi$ is $C^2$ near $b_{\mathrm{c}}$ and $\Phi(b')^+$ is locally bounded, $R_{\mathrm{nl}}(\Delta\theta) = \mathcal{O}(\|\Delta\theta\|^2)$ and $\Delta h_{\mathrm{comp}}(\Delta\theta)$ is well-defined near $b_{\mathrm{c}}$. Since every orthogonal projector satisfies $\|I - \Pi\|_2 \le 1$, we have
\begin{equation*}
\|\Delta G(\Delta\theta) - \delta G_{\mathrm{lin}}(\Delta\theta)\|_2 \le \|R_{\mathrm{nl}}(\Delta\theta)\|_2 = \mathcal{O}(\|\Delta\theta\|^2)
\end{equation*}
as $\Delta\theta \to 0$, which proves $\left.\frac{\partial(\Delta G)}{\partial\Delta\theta}\right|_{\Delta\theta=0} = \frac{\partial G_{\theta_{\mathrm{c}}}}{\partial \theta}$.
\end{proof}

\begin{proof}[Proof of Proposition~\ref{prop:stage2-descent}]
\emph{Part~(i): Bounded target update.} Write the singular value decomposition $\mathcal{R} = \sum_i \sigma_i u_i v_i^\top$ with $\sigma_i \ge 0$. For any damping $\mu > 0$, solving \eqref{eq:adjoint-cg} filters each singular mode explicitly:
\begin{equation*}
\begin{aligned}
\Delta a &= -\sum_i \frac{\sigma_i}{\sigma_i^2 + \mu}\langle u_i, r_{\mathrm{I}}\rangle v_i, \\
\mathcal{R}\,\Delta a &= -\sum_i \frac{\sigma_i^2}{\sigma_i^2 + \mu}\langle u_i, r_{\mathrm{I}}\rangle u_i.
\end{aligned}
\end{equation*}
For every $\sigma_i \ge 0$, the AM--GM inequality gives $\frac{\sigma_i}{\sigma_i^2 + \mu} \le \frac{1}{2\sqrt{\mu}}$ and $0 \le \frac{\sigma_i^2}{\sigma_i^2 + \mu} < 1$. Summing squared coefficients via Parseval's identity yields both bounds in \eqref{eq:response-gain}, independent of $\Gamma_{\max}$.

\emph{Part~(ii): First-order closed-loop descent.} Taking the inner product of \eqref{eq:adjoint-cg} with $\Delta a$ shows that the predicted residual shift always points opposite to $r_{\mathrm{I}}$:
\begin{equation}
\langle\mathcal{R}\Delta a,r_{\mathrm{I}}\rangle = -\|\mathcal{R}\Delta a\|_2^2 - \mu\|\Delta a\|_2^2.
\label{eq:app-normal-energy}
\end{equation}
Let $\mathcal{E} = \mathcal{J}_a - \mathcal{R}$ be the operator error, with $\|\mathcal{E}\Delta a\|_2 \le \epsilon\|\mathcal{J}_a\Delta a\|_2$ for $\epsilon \in (0, 1)$. By the reverse triangle inequality, $(1 - \epsilon)\|\mathcal{J}_a\Delta a\|_2 \le \|\mathcal{R}\Delta a\|_2$, so $\|\mathcal{E}\Delta a\|_2 \le \frac{\epsilon}{1-\epsilon}\|\mathcal{R}\Delta a\|_2$. Adding $\langle\mathcal{E}\Delta a, r_{\mathrm{I}}\rangle$ to \eqref{eq:app-normal-energy} and applying Cauchy--Schwarz yields $\langle\mathcal{J}_a\Delta a, r_{\mathrm{I}}\rangle \le -\mathcal{M}$, where
\begin{equation}
\mathcal{M} = \|\mathcal{R}\Delta a\|_2^2 + \mu\|\Delta a\|_2^2 - \frac{\epsilon}{1-\epsilon}\|\mathcal{R}\Delta a\|_2\|r_{\mathrm{I}}\|_2.
\label{eq:app-descent-margin}
\end{equation}
Whenever \eqref{eq:adjoint-descent} holds, $\mathcal{M} > \mu\|\Delta a\|_2^2 \ge 0$, so the directional derivative of $\mathcal{L}$ along $\Delta a$ is at most $-2\mathcal{M}/N < 0$.
\end{proof}

\paragraph{Remark (From output targets to parameter updates).}
Rescaling $\Delta a$ by a trust-region factor $c \in (0, 1]$ preserves its sign because the directional derivative is positively homogeneous: $\langle\mathcal{J}_a(c\Delta a), r_{\mathrm{I}}\rangle = c\langle\mathcal{J}_a\Delta a, r_{\mathrm{I}}\rangle \le -c\mathcal{M}$. When projecting $c\Delta a$ back onto all network weights via $\delta\theta = \theta_{\mathrm{proj}} - \theta^{\mathrm{I}}$, let $e_{\mathrm{cb}} = \frac{\partial G_{\theta^{\mathrm{I}}}}{\partial \theta}\,\delta\theta - \mathcal{P}(c\Delta a)$ be the callback fitting residual. Using $\mathcal{J}_a = \mathcal{J}_{\mathrm{cb}}\mathcal{P}$ gives
\begin{equation}
\begin{aligned}
\frac{N}{2}\left\langle\frac{\dd \mathcal{L}(\theta^{\mathrm{I}})}{\dd \theta},\delta\theta\right\rangle
&= \left\langle\mathcal{J}_{\mathrm{cb}} \frac{\partial G_{\theta^{\mathrm{I}}}}{\partial \theta}\,\delta\theta,r_{\mathrm{I}}\right\rangle \\
&= c\langle\mathcal{J}_a\Delta a, r_{\mathrm{I}}\rangle + \langle\mathcal{J}_{\mathrm{cb}} e_{\mathrm{cb}}, r_{\mathrm{I}}\rangle \\
&\le -c\mathcal{M} + \|\mathcal{J}_{\mathrm{cb}} e_{\mathrm{cb}}\|_2\|r_{\mathrm{I}}\|_2,
\end{aligned}
\label{eq:app-projected-descent}
\end{equation}
so weight projection preserves first-order closed-loop descent whenever $\|\mathcal{J}_{\mathrm{cb}} e_{\mathrm{cb}}\|_2\|r_{\mathrm{I}}\|_2 < c\mathcal{M}$.

\subsection{Stage I Implementation}
\label{app:stage1-details}

Stage~I updates the weights inside an $M$-dimensional subspace ($M \ll p$) to move the rollout into a locally smooth regime before Stage~II begins (Table~\ref{tab:app-hyperparams}).

\paragraph{Step 1: Choosing informative weight directions.}
Each round builds a small basis of $M \ll p$ joint readout--backbone directions $u_m$ spanning diverse output variations. In 2D, we combine leading readout directions from the cached penultimate features with layer-scaled random backbone directions and orthonormalize their output tangents via forward-mode derivatives on cached inputs (zero solver calls). In 3D POM, we draw $M = 8$ orthonormal mixed readout--backbone directions directly.

\paragraph{Step 2: Compensating finite-step nonlinearity.}
Before querying the solver at $\pm\rho_m u_m$, we apply low-rank readout compensation \eqref{eq:head-comp} to cancel the dominant higher-order output drift induced by nonlinear backbone layers (weighted over wet ocean cells in 3D POM).

\paragraph{Step 3: Solving and screening the Gauss--Newton update.}
From the $2M$ perturbed rollouts we form the central-difference sensitivity matrix $J$ and solve the regularized $M \times M$ Gauss--Newton system \eqref{eq:stage1-gn}, where $W$ damps directions exhibiting strong second-order curvature. We then apply readout compensation \eqref{eq:head-comp} to the combined step, evaluate a few candidate step sizes or damping values on forward rollouts, and keep the update with the lowest rollout loss (Table~\ref{tab:app-hyperparams}).

\subsection{Stage II Implementation}
\label{app:stage2-details}
\label{app:projection}

Whereas Stage~I moves along $M \ll p$ weight directions, Stage~II updates \emph{all} $p$ network weights around $\theta^{\mathrm{I}}$ without solver derivatives by learning how small perturbations to the closure's outputs propagate through the solver over time (Table~\ref{tab:app-hyperparams}).

\paragraph{Step 1: Output probing and held-out validation.}
Starting from the same initial states $q_0^{(w)}$, we inject $K$ paired output perturbations $\pm\delta a_t$ that are smooth in space and mutually orthogonal in time, and record the resulting state differences $d$ relative to the nominal rollout. In 3D POM, the perturbations are restricted to wet ocean cells and satisfy the same conservation constraints as the closure.

After fitting $\mathcal{R}$ on these probes, we evaluate its accuracy on a small disjoint set of held-out perturbations using the relative response error $\mathrm{RRE} = \|d - \mathcal{R}\,\delta a\|_2 / \|d\|_2$ and cosine similarity to gate the Stage~II parameter update.

\paragraph{Step 2: Structured causal response operator $\mathcal{R}$.}
Because the number of probes is far smaller than the grid dimension ($K \ll N$), fitting an unconstrained linear map is underdetermined. Instead, $\mathcal{R}$ uses a compact, physics-aligned linear parameterization matched to the domain geometry, making both forward products $\mathcal{R}\,\delta a$ and exact adjoint products $\mathcal{R}^\top y$ fast and well-conditioned.

\emph{Periodic 2D domains.} Across all four 2D benchmarks, state perturbations follow a linear time recurrence $d_{t+1} = A_t d_t + B_t\,\delta a_t$ ($d_0 = 0$). Each operator $A_t$ (and analogously $B_t$) combines a translation-invariant Fourier multiplier for spatially uniform wave propagation and diffusion with trajectory-dependent pointwise and first-order derivative modulations for spatially varying advection and interface coupling. We initialize the operators by one-step ridge regression, refine them across the multi-step window with L-BFGS, and cap the per-step operator gain to prevent multi-step blow-up. The exact adjoint $\mathcal{R}^\top y$ follows from the reverse-time recurrence $\zeta_T = y_T$, $\zeta_t = y_t + A_t^\top \zeta_{t+1}$ ($t = T-1, \dots, 1$), and $(\mathcal{R}^\top y)_t = B_t^\top \zeta_{t+1}$ ($t = 0, \dots, T-1$).

\emph{Bounded 3D ocean basin (POM).} Because the 3D ocean domain is a closed, non-periodic basin with zero-flux lateral walls and four hourly closure calls per window, $\mathcal{R}$ maps each hourly output perturbation causally to all subsequent scored states through compact block operators combining vertical column coupling on wet cells, zero-flux horizontal diffusion, and a few basin-scale spatial modes, rescaled so that $\|\mathcal{R}\|_2 \le \Gamma_{\max}$. Reusing the same probe rollouts at zero extra solver cost, we also fit a lightweight causal linear model predicting how earlier output perturbations shift the closure's later inputs $\delta x$, weighted by its held-out validation skill.

\paragraph{Step 3: Target inversion and full-network weight update.}
Once $\mathcal{R}$ is fitted, updating all $p$ weights requires no solver derivatives. We first solve the damped normal equation \eqref{eq:adjoint-cg} with trust-region conjugate gradients \citep{steihaug1983conjugate} to obtain step-wise target outputs $a_t^\star = a_t^{\mathrm{I}} + \Delta a_t$.

We then update all weights by matching the predicted closure outputs $\hat{a}_t = g_\theta(x_t^{\mathrm{I}})$ to $a_t^\star$ using standard backpropagation through $G_\theta$ alone (noting that $g_0(x_t^{\mathrm{I}})$ cancels identically in $\hat{a}_t - a_t^\star = G_\theta(x_t^{\mathrm{I}}) - (G_{\theta^{\mathrm{I}}}(x_t^{\mathrm{I}}) + \Delta a_t)$). In 2D, we minimize a discrete $H^2$ Sobolev regression loss on the cached inputs,
\begin{equation}
\begin{aligned}
\ell_{\mathrm{proj}}(\hat{a}, a^\star) &= \|\hat{a} - a^\star\|_2^2 + 0.1\|\nabla_{\mathrm{fd}}(\hat{a} - a^\star)\|_2^2 \\
&\quad + 0.01\|\Delta_{\mathrm{fd}}(\hat{a} - a^\star)\|_2^2,
\end{aligned}
\label{eq:app-sobolev}
\end{equation}
where $\nabla_{\mathrm{fd}}$ and $\Delta_{\mathrm{fd}}$ are unit-spacing periodic finite-difference gradient and Laplacian operators, alternating closed-form readout solves with Adam \citep{kingma2015adam} updates on the backbone \citep{golub1973differentiation}. In 3D POM, we initialize the linear readout from the target outputs via ridge regression on cached windows from recent rounds and then refine all weights with AdamW \citep{loshchilov2019decoupled} through the fitted response and input-shift operators. Finally, we evaluate a few candidate step scales on forward rollouts and keep the lowest-loss update (Table~\ref{tab:app-hyperparams}). Table~\ref{tab:app-ablation-map} maps each ablation in Table~\ref{tab:ablation} to these 2D and 3D components.

\begin{table*}[t]
\centering
\caption{\textbf{Summary of \ourmethod{} Stage~I and Stage~II hyperparameters across 2D and 3D benchmarks.}}
\label{tab:app-hyperparams}
\footnotesize
\renewcommand{\arraystretch}{1.08}
\setlength{\tabcolsep}{6pt}
\begin{tabularx}{\textwidth}{@{}>{\raggedright\arraybackslash}p{0.23\textwidth}>{\raggedright\arraybackslash}X>{\raggedright\arraybackslash}X@{}}
\toprule
Component / Hyperparameter & \textbf{2D Benchmarks} (CH, GS, NS, MHD) & \textbf{3D POM} \\
\midrule
Stage~I subspace \& rollouts & $M = 16$ ($8$ readout + $8$ backbone), output-orthonormal ($32$ batch probe evaluations) & $M = 8$ orthonormal mixed directions ($16$ batch probe evaluations) \\
Stage~I probe radius \& compensation & $\rho_m \le 0.02$; rank-$4$ readout compensation & $\rho = 0.35$; wet-weighted rank-$4$ readout compensation \\
Stage~I Gauss--Newton \& screening & Curvature-weighted $W$; $4$ dampings $\lambda \in [10^{-3}, 0.2]$ & Regularized GN (ridge $10^{-2}$); $3$ step scales $\alpha \in \{0.2, 0.5, 1.0\}$ \\
\midrule
Stage~II probes & $20$ fit + $4$ held-out subset evaluations ($\sigma = 0.02$, $32$-window subset) & $64$ fit + $16$ held-out window rollouts ($8$ fit + $2$ held-out per window; $\sigma = 0.20$) \\
Stage~II operator \& gain cap & Periodic recurrence + L-BFGS multi-step; step gain cap $10.5$ ($1.12$ on GS) & Causal hourly blocks ($\Gamma_{\max} = 4.0$) + causal input-shift model \\
Target CG \& weight update & Trust-region CG ($\mu = 0.01$, $\tau_a = \sigma$); $H^2$ projection \eqref{eq:app-sobolev}, $100$ Adam steps; $3$ scales & Trust-region CG ($\mu = 0.015$, $\tau_a = 0.25$); cached-window readout ridge init.\ + $60$ AdamW steps; $5$ candidates \\
\bottomrule
\end{tabularx}
\end{table*}

\begin{table*}[t]
\centering
\caption{\textbf{Component ablation settings for Table~\ref{tab:ablation}.}}
\label{tab:app-ablation-map}
\footnotesize
\renewcommand{\arraystretch}{1.08}
\setlength{\tabcolsep}{6pt}
\begin{tabularx}{\textwidth}{@{}>{\raggedright\arraybackslash}p{0.28\textwidth}>{\raggedright\arraybackslash}X>{\raggedright\arraybackslash}X@{}}
\toprule
Table~\ref{tab:ablation} Variant & \textbf{2D Benchmarks} & \textbf{3D POM} \\
\midrule
1.\ w/o Stage~I Warm-Up & Skip Stage~I ($\theta^{\mathrm{I}} = \theta_{\mathrm{c}}$); run Stage~II only & Skip Stage~I ($\theta^{\mathrm{I}} = \theta_{\mathrm{c}}$); run Stage~II only \\
2.\ w/o Stage~II Output Perturbation & Skip Stage~II ($\theta_{\mathrm{next}} = \theta^{\mathrm{I}}$); run Stage~I only & Skip Stage~II ($\theta_{\mathrm{next}} = \theta^{\mathrm{I}}$); run Stage~I only \\
3.\ Stage~I w/o Readout Comp. & Set $\Delta h_{\mathrm{comp}} = 0$ ($s = 0$) in Stage~I probes and step & Set $\Delta h_{\mathrm{comp}} = 0$ ($s = 0$) in Stage~I probes and step \\
4.\ Stage~II w/o State-Dep.\ Coupling & Drop state-dependent spatial modulations (Fourier multiplier only) & Disable causal input-shift feedback ($\delta x \equiv 0$) \\
5.\ Stage~II w/o Multi-Step Calib. & Use one-step ridge fit only (skip multi-step L-BFGS) & Restrict $\mathcal{R}$ to within-hour responses (drop cross-hour coupling) \\
6.\ Stage~II Head-Only Update & Freeze backbone $b = b^{\mathrm{I}}$; update readout $h$ only & Freeze backbone $b = b^{\mathrm{I}}$; update readout $h$ only \\
\bottomrule
\end{tabularx}
\end{table*}

\section{Benchmark Specifications}
\label{app:benchmarks}

Table~\ref{tab:app-benchmarks} summarizes the domains, resolutions, numerical integrators, closure architectures, and dataset splits across all five benchmarks. Throughout forward-only training of \ourmethod{} and all baselines, each simulator executes without access to solver derivatives or internal discretizations: the four 2D benchmarks run physical solvers in external CPU processes, while 3D NeuralPOM \citep{shu2026neuralpom} runs on GPU with autograd disabled so only forward states are returned under the basin geometry and conservation constraints \eqref{eq:app-pom-discrete}. For oracle reference, Solver BPTT \eqref{eq:bptt-gradient} differentiates a 64-bit PyTorch solver in 2D and differentiates through all $24$ internal solver steps in 3D.

In 2D, each state channel in $r_t(\theta) = q_t(\theta) - q_t^{\mathrm{ref}}$ is normalized by its training-set standard deviation; $G_\theta$ pads inputs with fewer than $8$ channels to $8$ at the lifting layer and scales normalized outputs by the training-set subgrid residual RMS $\sigma_{\mathrm{res}}$. 3D POM error normalizations for 4-hour MSE and 24-hour/48-hour NRMSE are detailed in Appendix~\ref{app:pom}. Physical parameters are defined locally within each subsection, with $T$ denoting ocean temperature in 3D POM equations and figures.

\begin{table*}[!htb]
\centering
\caption{\textbf{Summary of the five hybrid PDE benchmarks.} Independent reference trajectories are partitioned into disjoint training, validation, and test splits before extracting rollout windows ($N_{\mathrm{cb}}$ denotes closure calls per window, In $\to$ out ch.\ gives input and output channels of $G_\theta$, and RA denotes the Robert--Asselin filter). On 3D POM, each 4-hour window executes $24$ internal $600\text{ s}$ steps ($12$ twenty-minute scored outputs) with $N_{\mathrm{cb}} = 4$ hourly closure calls.}
\label{tab:app-benchmarks}
\footnotesize
\renewcommand{\arraystretch}{1.12}
\setlength{\tabcolsep}{3pt}
\begin{tabularx}{\textwidth}{@{}>{\raggedright\arraybackslash}p{0.20\textwidth}>{\raggedright\arraybackslash}p{0.16\textwidth}>{\raggedright\arraybackslash}p{0.27\textwidth}>{\raggedright\arraybackslash}X@{}}
\toprule
Benchmark & Domain & Reference $\to$ Coarse & Coarse numerical solver ($H$) \\
\midrule
2D Cahn--Hilliard & $[0,1.0)^2$ & $384^2 \to 96^2$ ($4\times$ avg.) & Stabilized IMEX Fourier Euler \\
2D Gray--Scott & $[0,2.5)^2$ & $768^2 \to 96^2$ ($8\times$ avg.) & Strang exact diffusion + midpoint reaction \\
2D Kolmogorov NS & $[0,2\pi)^2$ & $256^2 \to 64^2$ (spectral trunc.) & 4-stage Cox--Matthews ETD-RK4 \\
2D Guide-Field MHD & $[0,2\pi)^2$ & $256^2 \to 64^2$ (spectral trunc.) & Coupled $2\times 2$ matrix ETD-RK4 \\
\addlinespace[2pt]
3D Ocean Model (POM) & \begin{tabular}[t]{@{}l@{}}$0^\circ\text{--}22^\circ\text{E}$,\\$30^\circ\text{--}50^\circ\text{N}$\end{tabular} & \begin{tabular}[t]{@{}l@{}}$354\!\times\!322\!\times\!16\,\sigma \to$\\$90\!\times\!82\!\times\!16\,\sigma$ ($4\times$ avg.)\end{tabular} & Semi-implicit Helmholtz CG + Thomas + RA \\
\bottomrule
\end{tabularx}
\par\vspace{5pt}
\begin{tabular*}{\textwidth}{@{\extracolsep{\fill}}llcrr@{}}
\toprule
Benchmark & Network $G_\theta$ (params) & In $\to$ out ch. & $N_{\mathrm{cb}}$ & Train / Val / Test windows \\
\midrule
2D Cahn--Hilliard & 4-layer FNO ($489{,}057$) & $2 \to 1$ & $100$ & $2{,}352$ / $784$ / $784$ \\
2D Gray--Scott & 4-layer FNO ($489{,}057$) & $4 \to 1$ & $480$ & $4{,}752$ / $1{,}584$ / $1{,}584$ \\
2D Kolmogorov NS & Full-spectrum FNO ($490{,}027$) & $6 \to 2$ & $160$ & $360$ / $120$ / $120$ \\
2D Guide-Field MHD & Full-spectrum FNO ($490{,}057$) & $8 \to 3$ & $160$ & $360$ / $120$ / $120$ \\
3D Ocean Model (POM) & 8-block ResNet ($4{,}271{,}172$) & $161 \to 60$ & $4$ & $1{,}800$ / $16$ / $48$ \\
\bottomrule
\end{tabular*}
\end{table*}

\subsection{2D Cahn--Hilliard}
\label{app:ch}

\paragraph{Governing PDE and reference data.}
Scalar concentration $\phi(\mathbf{x},t)$ on $\Omega = [0, L)^2$ ($L = 1.0$, domain mean $\bar{\phi} = -0.1$) evolves under fourth-order conserved spinodal decomposition \citep{cahn1958free,cahn1961spinodal}:
\begin{equation}
\partial_t \phi = \mu_{\mathrm{CH}}\Delta\big(\phi^3 - \phi - \kappa\Delta\phi\big),
\label{eq:app-ch-pde}
\end{equation}
with mobility $\mu_{\mathrm{CH}} = 10^{-3}$ and gradient-energy coefficient $\kappa = 1.5625 \times 10^{-4}$ ($\sqrt{\kappa} = 0.0125$). Both fine ($384 \times 384$) and coarse ($96 \times 96$) meshes discretize spatial derivatives with the periodic five-point finite-volume Laplacian $\Delta_h$, whose discrete Fourier symbol on grid spacing $h_x = L/N_g$ ($N_g$ grid points per axis) is
\begin{equation}
\widehat{\Delta}_h(\mathbf{k}) = -\frac{4}{h_x^2}\!\left[\sin^2\!\left(\frac{k_x h_x}{2}\right) + \sin^2\!\left(\frac{k_y h_x}{2}\right)\right].
\label{eq:app-discrete-lap}
\end{equation}
Fine-grid simulations are $4 \times 4$ conservatively block-averaged onto $96 \times 96$ across $48$ training, $16$ validation, and $16$ test trajectories ($500$ output steps of $\Delta t_{\mathrm{out}} = 0.1$; stride-$10$ extraction yields $49$ twenty-step windows per trajectory).

\paragraph{Coarse hybrid step and shared initialization.}
Each coarse interval takes five substeps ($\Delta t = 0.02$, $N_{\mathrm{cb}} = 100$). At substep $n$, $F(\phi^n) = (\phi^n, \kappa\Delta_h\phi^n) \in \R^{2\times 96\times 96}$, and a four-layer periodic FNO \citep{li2021fourier} $G_\theta$ (width $29$, $6$ Fourier modes per axis, $489{,}057$ parameters) predicts the bulk chemical potential $a^n = (\phi^n)^3 + G_\theta(x^n)$. The time stepper $H(\phi^n, a^n)$ advances $\phi^n$ via stabilized semi-implicit Fourier Euler ($S_{\mathrm{stab}} = 2.0$):
\begin{equation}
\begin{aligned}
\Big(1 &- \Delta t\,\mu_{\mathrm{CH}} S_{\mathrm{stab}}\widehat{\Delta}_h + \Delta t\,\mu_{\mathrm{CH}}\kappa\widehat{\Delta}_h^2\Big)\widehat{\phi}^{n+1} \\
&= \widehat{\phi}^n + \Delta t\,\mu_{\mathrm{CH}}\widehat{\Delta}_h\mathcal{F}\!\big(a^n - (1 + S_{\mathrm{stab}})\phi^n\big),
\end{aligned}
\label{eq:app-ch-imex}
\end{equation}
where $\mathcal{F}$ is the 2D discrete Fourier transform. Since $\widehat{\Delta}_h(\mathbf{0}) = 0$, $H$ conserves $\sum_{i,j}\phi_{i,j}^n$ to machine precision for any $a^n$.

On Cahn--Hilliard, all methods start from a shared offline-pretrained checkpoint $\theta_0$, evaluating closed-loop refinement of an offline closure alongside the zero-output cold starts used on the other four benchmarks. We pretrain $\theta_0$ offline on $24{,}000$ input--output pairs from the training split using the discrete $H^2$ loss \eqref{eq:app-sobolev} (batch size $8$, Adam with $\mathrm{lr} = 10^{-4}$, up to $64$ epochs, selected at epoch $27$), yielding a closed-loop validation MSE of $2.760 \times 10^{-4}$ (static paired validation MSE $7.505 \times 10^{-3}$). Cold-start ($h_0 = 0$) results on Cahn--Hilliard appear in Appendix~\ref{app:ch-coldstart}.

\subsection{2D Gray--Scott}
\label{app:gs}

\paragraph{Governing PDE and reference data.}
Two chemical species $(u, v)$ on $[0, L)^2$ ($L = 2.5$) evolve under autocatalytic reaction--diffusion dynamics \citep{pearson1993complex}:
\begin{equation}
\begin{aligned}
\partial_t u &= D_u\Delta u - u v^2 + f(1 - u), \\
\partial_t v &= D_v\Delta v + u v^2 - (f + k_r)v,
\end{aligned}
\label{eq:app-gs-pde}
\end{equation}
with diffusivities $(D_u, D_v) = (2\times 10^{-5}, 1\times 10^{-5})$, feed rate $f = 0.035$, and kill rate $k_r = 0.062$. Reference simulations on $768 \times 768$ are $8 \times 8$ block-averaged onto $96 \times 96$ across $48$ training, $16$ validation, and $16$ test trajectories ($1{,}000$ intervals of $\Delta t_{\mathrm{out}} = 10.0$, giving $99$ twenty-step windows per trajectory at stride $10$).

\paragraph{Coarse hybrid step.}
Each output interval executes $12$ Strang-split substeps ($\Delta t = 10/12$), bracketing a two-stage explicit midpoint reaction step with exact half-step Fourier diffusion $\exp(\frac{\Delta t}{2}D_{u,v}\widehat{\Delta}_h)$ ($N_{\mathrm{cb}} = 480$). At each stage, $F(u,v) = (u, v, D_u\Delta_h u/0.05, D_v\Delta_h v/0.05) \in \R^{4\times 96\times 96}$, and the four-layer periodic FNO ($489{,}057$ parameters, $h_0 = 0$) predicts the effective reaction rate $a = u v^2 + G_\theta(x)$. Opposite-sign stoichiometric increments $(\mp a)$ in $H$ preserve exact reactant conversion between $u$ and $v$.

\subsection{2D Kolmogorov NS}
\label{app:ns}

\paragraph{Governing PDE and reference data.}
Incompressible vorticity $\omega = \partial_x v - \partial_y u$ of velocity $\mathbf{u} = (u, v)$ with streamfunction $\psi$ on $[0, 2\pi)^2$ evolves under monochromatic shear forcing and linear Ekman drag \citep{chandler2013invariant,kochkov2021machine}:
\begin{equation}
\begin{aligned}
\partial_t \omega + \nabla\cdot(\mathbf{u}\omega) &= \nu\Delta\omega - \alpha\omega - A k_f\cos(k_f y), \\
\Delta\psi &= \omega, \qquad \mathbf{u} = (-\partial_y\psi,\, \partial_x\psi),
\end{aligned}
\label{eq:app-ns-pde}
\end{equation}
with viscosity $\nu = 5\times 10^{-4}$, drag $\alpha = 0.1$, forcing amplitude $A = 1.0$, wavenumber $k_f = 6$, and $25.0$ time units of spinup ($\widehat{\psi}(\mathbf{k}) = -\widehat{\omega}(\mathbf{k})/|\mathbf{k}|^2$ for $\mathbf{k} \neq \mathbf{0}$). Reference $256 \times 256$ trajectories ($10$ substeps per $\Delta t_{\mathrm{out}} = 0.02$) are spectrally truncated with zeroed Nyquist modes onto $64 \times 64$ across $24$ training, $8$ validation, and $8$ test trajectories ($160$ intervals and $15$ twenty-step windows per trajectory).

\paragraph{Coarse hybrid step.}
The coarse $64 \times 64$ solver runs two substeps ($\Delta t = 0.01$) of four-stage Cox--Matthews ETD-RK4 \citep{cox2002exponential} per interval ($N_{\mathrm{cb}} = 160$). For linear symbol $z(\mathbf{k}) = -\Delta t(\nu|\mathbf{k}|^2 + \alpha)$, stage coefficients $(\mathrm{e}^{z/2}, \mathrm{e}^z, \varphi_1, \varphi_2, \varphi_3)$ are precomputed in 64-bit precision via $4\times 4$ block matrix exponentials. At each stage, $F(\omega) = (\omega, \psi, u, v, \partial_x\omega, \partial_y\omega) \in \R^{6\times 64\times 64}$ and $g_0(x) = (u\odot_{3/2}\omega,\, v\odot_{3/2}\omega)$ is the $3/2$-dealiased advective vorticity flux. A full-spectrum periodic FNO ($490{,}027$ parameters, $h_0 = 0$) combines $5$ low-wavenumber mode weights, $14$ radial shells, a $16$-channel $2\times$-dealiased bilinear branch, and a $6$-band spectral readout; $a = g_0(x) + G_\theta(x)$ enters $H$ via spectral divergence $-i\mathbf{k}\cdot\widehat{a}(\mathbf{k})$, preserving zero mean vorticity identically.

\newcommand{\pomWideEquations}{%
\begin{wideblock}
{\small
\begin{equation}
\left\{
\begin{aligned}
&\partial_t \eta + \nabla_h \cdot (D\bar{\mathbf{u}}_h) = 0, \qquad
  \partial_\sigma \omega_\sigma = -\nabla_h \cdot (D\mathbf{u}_h) - \partial_t \eta \quad (\omega_\sigma|_{\sigma=0,-1} = 0), \\[2pt]
&\partial_t(D\mathbf{u}_h) + \nabla_h \cdot (D\mathbf{u}_h \otimes \mathbf{u}_h) + \partial_\sigma(\omega_\sigma \mathbf{u}_h) + f\hat{\mathbf{k}} \times D\mathbf{u}_h
  = -g D \nabla_h \eta - \frac{g D}{\rho_0}\bm{\mathcal{B}}_\rho(\sigma) + \bm{\mathcal{F}}_h(\mathbf{u}_h) + \partial_\sigma\!\left(\frac{K_M}{D}\partial_\sigma \mathbf{u}_h\right), \\[2pt]
&\partial_t(D\chi) + \nabla_h \cdot (D\mathbf{u}_h \chi) + \partial_\sigma(\omega_\sigma \chi)
  = \nabla_h \cdot (D A_H \nabla_h \chi) + \partial_\sigma\!\left(\frac{K_H}{D}\partial_\sigma \chi\right) + \delta_{\chi T}\frac{\partial_\sigma I_{\mathrm{sw}}(\sigma)}{\rho_0 c_p}, \qquad \chi \in \{T, S\}.
\end{aligned}
\right.
\label{eq:app-pom-continuous}
\end{equation}
\vspace{2pt}
\begin{equation}
\left\{
\begin{aligned}
&\text{(Helmholtz CG):} \;
  \Big(I - \alpha_{\mathrm{fs}} g (2\Delta t_{\mathrm{3D}})^2 \nabla_h \cdot D^n \nabla_h\Big)\eta^{n+1} = \mathcal{S}_{\mathrm{ext}}\!\big(\eta^{n-1}, \bar{\mathbf{u}}_h^{\,\star}, \bm{\mathcal{G}}_{\mathrm{bc}}^n\big), \; \bar{\mathbf{u}}_h^{n+1} = \bm{\mathcal{C}}_{\mathrm{bt}}\!\big(\bar{\mathbf{u}}_h^{\,\star}, \nabla_h\eta^{n+1}\big), \\[2pt]
&\text{(Predictor \& hourly $F$):} \;
  \bm{\psi}^{\mathrm{prov}} = \bm{\psi}^{n-1} + 2\Delta t_{\mathrm{3D}}\,\bm{\mathcal{T}}_{\mathrm{expl}}(q^{n-1}, q^n), \; x^{6m} = F(q^{6m-1}, q^{6m}) \in \R^{161\times 90\times 82}, \\[2pt]
&\text{(Conservative closure):} \;
  a^{\mathrm{hr},\,m} = G_{\theta}(x^{6m}) \in \R^{60\times 90\times 82}, \;\;
  \sum_{l=1}^{15} \Delta\sigma_l\,a_{u,v,l}^{\mathrm{hr},\,m} = 0, \;\; \sum_{(i,j)\,\mathrm{wet}} \!\!H_{\mathrm{b},i,j} A_{i,j}\,a_{T,S,l}^{\mathrm{hr},\,m} = 0, \\[2pt]
&\text{(Implicit vertical $H$):} \;
  \left(I - \frac{2\Delta t_{\mathrm{3D}}}{(D^{n+1})^2}\partial_\sigma K_{M,H}^n\partial_\sigma\right)\bm{\psi}^{n+1} = \bm{\psi}^{\mathrm{prov}} + \frac{1}{6}\,a^{\mathrm{hr},\,m} \quad (m = \lfloor n/6\rfloor).
\end{aligned}
\right.
\label{eq:app-pom-discrete}
\end{equation}
}
\end{wideblock}%
}

\makeatletter\if@twocolumn\addtocounter{equation}{2}\pomWideEquations\addtocounter{equation}{-4}\fi\makeatother

\subsection{2D Guide-Field MHD}
\label{app:mhd}

\paragraph{Governing PDE and reference data.}
We evolve coupled incompressible vorticity $\omega = \Delta\psi$ and out-of-plane magnetic potential $\psi_m$ on $[0, 2\pi)^2$, with velocity $\mathbf{u} = (-\partial_y\psi, \partial_x\psi)$, uniform guide field $\mathbf{B}_0 = B_0\mathbf{e}_x$ ($B_0 = 0.5$), fluctuating magnetic field $\mathbf{b} = (-\partial_y\psi_m, \partial_x\psi_m)$, and current density $j = \Delta\psi_m$. The governing equations \citep{biskamp2003magnetohydrodynamic} are
\begin{equation}
\begin{aligned}
\partial_t \omega + \nabla\cdot(\mathbf{u}\omega - \mathbf{b}j)
  &= B_0\partial_x j + \nu\Delta\omega \\
  &\quad - \alpha\omega + f_\omega, \\
\partial_t \psi_m + \mathbf{u}\cdot\nabla\psi_m
  &= B_0\partial_x \psi + \eta_m\Delta\psi_m,
\end{aligned}
\label{eq:app-mhd-pde}
\end{equation}
with $\nu = \eta_m = 5\times 10^{-4}$, drag $\alpha = 0.05$, cellular forcing $f_\omega = -A k_f\cos(k_f x)\cos(k_f y)$ ($A = 1.0$, $k_f = 4$), and $35.0$ time units of spinup. Reference $256 \times 256$ trajectories ($\Delta t_{\mathrm{out}} = 0.025$) are spectrally truncated onto $64 \times 64$ across $24$ training, $8$ validation, and $8$ test trajectories ($160$ intervals and $15$ twenty-step windows per trajectory at stride $10$).

\paragraph{Coarse hybrid step.}
Under $\partial_x \leftrightarrow +i k_x$ with $\widehat{\psi}(\mathbf{k}) = -\widehat{\omega}(\mathbf{k})/|\mathbf{k}|^2$ and $\widehat{j}(\mathbf{k}) = -|\mathbf{k}|^2\widehat{\psi}_m(\mathbf{k})$, linear Alfv{\'e}n wave coupling and dissipation form a $2\times 2$ non-normal Fourier symbol for $\mathbf{k} \neq \mathbf{0}$:
\begin{equation}
L(\mathbf{k}) = \begin{pmatrix} -\nu|\mathbf{k}|^2 - \alpha & -i k_x B_0 |\mathbf{k}|^2 \\ -i k_x B_0 / |\mathbf{k}|^2 & -\eta_m |\mathbf{k}|^2 \end{pmatrix},
\label{eq:app-mhd-matrix}
\end{equation}
integrated via $8\times 8$ block matrix exponentials in four-stage matrix ETD-RK4 ($2$ substeps of $\Delta t = 0.0125$, $N_{\mathrm{cb}} = 160$). At each stage, $F(\omega, \psi_m) = (\omega, \psi_m, \psi, j, u, v, b_x, b_y) \in \R^{8\times 64\times 64}$, and the full-spectrum periodic FNO ($490{,}057$ parameters, $h_0 = 0$) corrects the three-channel Reynolds--Maxwell stress and electromotive force $g_0(x) = (u\odot_{3/2}\omega - b_x\odot_{3/2}j,\, v\odot_{3/2}\omega - b_y\odot_{3/2}j,\, u\odot_{3/2}b_y - v\odot_{3/2}b_x)$ before spectral divergence is applied to the vorticity--current flux and the electromotive force enters the induction equation.
\makeatletter\if@twocolumn\addtocounter{equation}{2}\fi\makeatother

\subsection{3D Ocean Circulation (POM)}
\label{app:pom}

\paragraph{Governing equations.}
The 3D benchmark embeds a baroclinic closure inside the semi-implicit free-surface ocean solver NeuralPOM \citep{shu2026neuralpom}, discretizing the hydrostatic $\sigma$-coordinate primitive equations of the Princeton Ocean Model \citep{blumberg1987description,xu2015pomgpu} on a closed rectangular double-gyre basin of $(N_x, N_y, N_z) = (90, 82, 16)$ grid points ($0.25^\circ$ spherical grid spanning $0^\circ$--$22^\circ\text{E}$ and $30^\circ$--$50^\circ\text{N}$, $15$ uniform $\sigma$-layers, flat resting depth $H_{\mathrm{b}} = 2{,}500\text{ m}$, and zero-flux lateral walls; Fig.~\ref{fig:app-ocean}a). With $\sigma = (z - \eta)/D \in [-1, 0]$, column thickness $D = H_{\mathrm{b}} + \eta$, horizontal velocity $\mathbf{u}_h = (u,v)^\top$, depth-averaged velocity $\bar{\mathbf{u}}_h = \int_{-1}^0 \mathbf{u}_h\,\dd\sigma$, cross-$\sigma$ velocity $\omega_\sigma$, temperature $T$, and salinity $S$, the core primitive equations are given in \eqref{eq:app-pom-continuous}.
\makeatletter\if@twocolumn\else\pomWideEquations\fi\makeatother
Here $f$ is the Coriolis parameter, $g$ gravity, $\rho_0$ reference density, $c_p$ seawater specific heat, and $I_{\mathrm{sw}}$ penetrative shortwave irradiance. Density anomaly $\rho' = \rho_{\mathrm{EOS}}(T,S,z) - \rho_{\mathrm{ref}}(\sigma)$ \citep{mellor1991equation} determines the baroclinic pressure gradient $\bm{\mathcal{B}}_\rho(\sigma) = D\int_\sigma^0 \nabla_h\rho'\,\dd\sigma' - \nabla_h D\int_\sigma^0 \sigma'\partial_{\sigma'}\rho'\,\dd\sigma'$. Horizontal viscous stress divergence $\bm{\mathcal{F}}_h$ and tracer mixing $A_H$ follow the Smagorinsky scheme \citep{smagorinsky1963general}, while vertical diffusivities $(K_M, K_H)$ follow the Mellor--Yamada $2.5$ turbulence closure \citep{mellor1982development,blumberg1987description}.

\paragraph{Semi-implicit hybrid step.}
On an Arakawa C-grid with leapfrog internal step $\Delta t_{\mathrm{3D}} = 600\text{ s}$ ($24$ internal steps per 4-hour window), each step $n$ solves \eqref{eq:app-pom-discrete}: a spectral-preconditioned CG Helmholtz solve ($\alpha_{\mathrm{fs}} = 0.333$, relative tolerance $10^{-5}$, up to $200$ iterations, with external right-hand side $\mathcal{S}_{\mathrm{ext}}$ incorporating the baroclinic forcing $\bm{\mathcal{G}}_{\mathrm{bc}}^n$) updates free-surface elevation $\eta^{n+1}$ and barotropic velocity $\bar{\mathbf{u}}_h^{n+1}$ via the barotropic correction $\bm{\mathcal{C}}_{\mathrm{bt}}$; the hourly closure tendency $a^{\mathrm{hr},\,m}$ ($m = \lfloor n/6\rfloor \in \{0,1,2,3\}$) is injected into the stacked $15$-layer baroclinic state $\bm{\psi} = (u, v, T, S)$ prior to tridiagonal Thomas implicit vertical diffusion; and a mode-preserving Robert--Asselin (RA) filter \citep{asselin1972frequency} ($\alpha_{\mathrm{RA}} = 0.15$) damps leapfrog computational modes while preserving depth-averaged velocity. Once per hour ($N_{\mathrm{cb}} = 4$), $F$ exports $161$ normalized channels combining $101$ current/lagged state, diffusivity, forcing, and geometry fields with $60$ explicit predictor increments. An eight-block dilated ResNet ($161 \to 256 \to 60$; $4{,}271{,}172$ parameters including $15{,}420$ readout weights $h$, initialized at $h_0 = 0$) outputs $a^{\mathrm{hr},\,m}$, masking dry cells and enforcing the zero column-mean momentum and zero volume-weighted tracer constraints in \eqref{eq:app-pom-discrete} (with horizontal cell areas $A_{i,j}$).

\paragraph{Reference ensemble and splits.}
Reference trajectories come from an eight-member, $9{,}000$-day ensemble ($1{,}000$-day spinup) of a $4\times$ horizontally refined ($354\times 322\times 16\,\sigma$, $0.0625^\circ$) eddy-resolving configuration, coarse-grained onto $(90, 82, 16)$ via wet-area cell averaging for $(T, S, \eta)$ and face-length flux averaging for $(u, v)$. Six training members (days $1{,}000$--$6{,}980$, $20$-day stride) provide $1{,}800$ 4-hour windows ($8$ windows per round); a separate validation member provides $16$ 4-hour windows (days $7{,}000$--$7{,}900$) plus eight 48-hour rollouts from a 9-day continuation; and a held-out test member provides $48$ rollouts (days $8{,}000$--$8{,}940$, $20$-day stride) for both 4-hour test MSE and 48-hour forecasts.

\paragraph{Evaluation metrics.}
The 4-hour trajectory MSE averages wet-volume-weighted squared errors across $(u, v, T, S, \eta)$ (area-weighted for $\eta$) over the $12$ twenty-minute scored steps, normalized per field by the unparameterized coarse solver's 4-hour validation RMS error. Instantaneous 24-hour and 48-hour composite NRMSE averages each field's wet-weighted RMSE normalized by its reference climatological standard deviation $\sigma_f$ across $(u, v, T, S, \eta)$, and 48-hour surface kinetic energy (KE) relative spectral error computes $\frac{1}{K_{\max}}\sum_{k=1}^{K_{\max}} |E_K(k)/E_K^{\mathrm{ref}}(k) - 1|$ ($K_{\max} = 31$) on the 48-trajectory mean spectrum.

\section{Baseline Protocols and Query Accounting}
\label{app:baselines}

\subsection{Shared Protocol, Hardware, and Stopping Rules}

All forward-only methods start from the same checkpoint $\theta_0$, draw from the same training window pools, and are evaluated deterministically (without exploration noise) on identical validation and test splits. Each training run uses a single fixed seed (seed~$0$); reported means and standard deviations measure spread across the held-out test trajectories (Table~\ref{tab:app-benchmarks}). On the four 2D benchmarks, training runs up to $100$ rounds (or matched window budget for Stage-I-Only) with full-split validation every $2$ rounds and stops early after $5$ checks without a $\ge 0.2\%$ relative MSE drop (neural updates run on GPU paired with external CPU solver workers across multi-GPU H100/A100 nodes). On 3D POM, all forward-only methods share a $121{,}504$-window budget cap on $8\times\text{A100}$ ($80\text{ GB}$) GPUs with $16$-window validation every round and early stopping after $20$ rounds without an improvement exceeding $10^{-6}$. Final checkpoints are selected by lowest validation MSE.

\subsection{Baseline Configurations}

Each baseline family is tuned and run under the shared forward-only protocol as follows:
\begin{itemize}[leftmargin=1.3em,itemsep=2pt,topsep=2pt]
  \item \textbf{ES} \citep{salimans2017evolution}: Antithetic pairs in a $16$-D QR-orthonormal readout--backbone subspace, using $24$ evaluations per round at $\sigma_{\mathrm{param}} = 10^{-3}$ with a 7-point line search in 2D and $16$ evaluations per round at $\sigma_{\mathrm{param}} = 0.08$ with a 4-point line search in 3D.
  \item \textbf{EKI} \citep{kovachki2019ensemble,lopezgomez2022training}: A $24$-member ensemble in the same $16$-D subspace, using $\sigma_{\mathrm{param}} = 10^{-3}$ and trace-relative ridge $0.01\,\mathrm{tr}(G)/16$ (where $G$ is the empirical residual Gram matrix) with a 6-point line search in 2D, and $\sigma_{\mathrm{param}} = 0.08$ with Tikhonov regularization $0.05$ and a 4-point line search in 3D.
  \item \textbf{Global PPO and Local PPO} \citep{schulman2017proximal,novati2021automating}: Clipped policy-gradient updates ($\epsilon_{\mathrm{clip}} = 0.2$, $4$ epochs per round, exploration $\sigma_a = 0.05$, target $\mathrm{KL} = 0.02$ in 2D and $0.05$ in 3D). In 2D, Global and Local PPO use trajectory-scalar and grid-local Monte Carlo returns relative to the nominal rollout with a learned spatial critic and forward rollout screening ($\mathrm{lr} = 10^{-4}$ on Cahn--Hilliard; tuned in $[10^{-5}, 10^{-4}]$ on the other 2D benchmarks). On 3D POM, a single clipped PPO baseline uses standardized window-loss reductions relative to the nominal rollout with KL backtracking ($\mathrm{lr} = 5\times 10^{-7}$).
  \item \textbf{Closure-RL} \citep{bassewitz2025closure}: Spatial grid-wise policy gradients ($\sigma_a = 0.05$, target spatial $\mathrm{KL} = 0.02$, actor learning rates in $[10^{-5}, 10^{-4}]$ in 2D and $2\times 10^{-5}$ in 3D) with 64-bit actor precision in 2D and 64-bit ratio/KL statistics in 3D.
  \item \textbf{Neural Emulator Method} \citep{frezat2026gradient}: Alternating hybrid rollout collection and multi-step autoregressive solver-emulator training to backpropagate updates into $G_\theta$, with validation and test rollouts evaluated inside the true coarse solver. We use a $3$-block width-$32$ FNO ($60$ emulator and $20$ closure steps per round) in 2D and a $4$-block width-$96$ dilated ResNet ($40$ emulator and $12$ closure steps per round) in 3D at $\mathrm{lr} = 3\times 10^{-4}$, followed by solver screening over step scales $\alpha \in \{1, 0.5, 0.25\}$.
  \item \textbf{Solver BPTT} \citep{werbos1990backpropagation,um2020solver}: End-to-end differentiation through 64-bit PyTorch solvers up to $128$ epochs in 2D with cosine learning-rate schedules ($\mathrm{lr} = 3\times 10^{-5}$ on Cahn--Hilliard and Gray--Scott; $\mathrm{lr} = 3\times 10^{-3}$ on Kolmogorov NS and MHD) and across all $24$ semi-implicit steps up to $300$ epochs in 3D using AdamW ($\text{weight decay } 10^{-4}$, gradient clipping, and plateau learning-rate decay, selected by validation MSE at epoch $298$).
\end{itemize}

On 3D POM, random parameter subspaces without output orthonormalization produce empirical Gram eigenvalues seven orders of magnitude below the regularization floor, explaining why weight-space zeroth-order methods barely move. On 2D Kolmogorov NS and MHD, scalar trajectory rewards in Global PPO dilute credit assignment across more than $10^6$ spatiotemporal action degrees of freedom.

\subsection{Query Accounting and Euler Approximations}
\label{app:compute}
\label{app:ega}

\paragraph{Rollout accounting.}
In Fig.~\ref{fig:efficiency}, cumulative rollout counts sum every training-split window evaluation across probing and step-size screening while excluding periodic validation sweeps. In Fig.~\ref{fig:efficiency}d, 3D POM GPU-hours measure total wall-clock time (forward solver rollouts plus neural updates) on matched $8\times\text{A100}$ nodes.

\paragraph{Inapplicability of Euler gradient approximations (EGA).}
For additive state corrections $q_{t+1} = q_t + \Delta t(f_c(q_t) + a_t)$, \citet{ouala2024online} formulate online Euler gradient approximations (EGA) with static, tangent-linear, and ensemble state-propagation variants. None applies directly to our forward-only embedded closure setting: the tangent-linear variant requires solver derivatives $\partial \mathcal{M}/\partial q$ unavailable under forward-only execution, and the ensemble variant targets additive state tendencies rather than an embedded $a_t \mapsto H(q_t, a_t)$ interface on high-dimensional PDE grids. Finally, the static approximation $\frac{\partial q_{t+1}}{\partial a_t} \approx \Delta t\,I$ assumes $a_t$ acts as a direct additive update on every state variable, ignoring spatial operators ($\mu_{\mathrm{CH}}\Delta$ in Cahn--Hilliard, $-\nabla\cdot$ in Kolmogorov NS and MHD), conservation projections, and cases where $a_t$ and $q_t$ differ in dimension (such as a single reaction rate updating two species in Gray--Scott, or baroclinic tendencies excluding free-surface elevation $\eta$ in 3D POM). Applying this static approximation on 2D Cahn--Hilliard---where the chemical potential enters \eqref{eq:app-ch-imex} through the discrete Laplacian $\mu_{\mathrm{CH}}\widehat{\Delta}_h(\mathbf{k})$ and semi-implicit denominator rather than the identity---causes training to stall at $1.94\times 10^{-4}$ validation MSE, compared with $4.06\times 10^{-5}$ for \ourmethod{}.

\section{Additional Experimental Results}
\label{app:extended-results}

\subsection{Oracle Verification and Cold-Start Cahn--Hilliard}
\label{app:ch-coldstart}

Table~\ref{tab:app-oracle-2d} reports the converged white-box Solver BPTT test errors on the four 2D benchmarks that define the Oracle Recovery row in Table~\ref{tab:main-2d}, confirming that \ourmethod{} recovers $87.7\%$--$97.1\%$ of the white-box error reduction without solver derivatives.

Table~\ref{tab:app-ch-coldstart} evaluates 2D Cahn--Hilliard from a zero-output cold start ($h_0 = 0$) rather than $\theta_0$. Because the mobility Laplacian $\mu_{\mathrm{CH}}\Delta$ and biharmonic damping $\kappa\Delta^2$ strongly attenuate small fixed-scale parameter perturbations, ES, EKI, Global/Local PPO, and the Neural Emulator Method remain near Coarse Physics, whereas calibrating perturbations in output space enables \ourmethod{} to cut cold-start test MSE by $63\%$.

\begin{table*}[t]
\centering
\caption{\textbf{2D white-box Solver BPTT test verification and Oracle Recovery.} Mean $\pm$ s.d.\ test MSE across independent test trajectories alongside the validation-selected BPTT epoch; Oracle Recovery computes $(\mathrm{MSE}_{\mathrm{Coarse}} - \mathrm{MSE}_{\mathrm{FORGE}})/(\mathrm{MSE}_{\mathrm{Coarse}} - \mathrm{MSE}_{\mathrm{BPTT}})$ from unrounded means, matching Table~\ref{tab:main-2d}.}
\label{tab:app-oracle-2d}
\fontsize{8}{9.5}\selectfont
\renewcommand{\arraystretch}{1.08}
\setlength{\tabcolsep}{6pt}
\begin{tabularx}{\textwidth}{@{}>{\raggedright\arraybackslash}Xcccc@{}}
\toprule
Benchmark (scale) & Coarse Physics & \textbf{\ourmethod{} (Ours)} & \emph{Solver BPTT (oracle, epoch)} & \emph{Oracle Recovery} \\
\midrule
2D Cahn--Hilliard ($\times 10^{-5}$) & $45.2$\sd{4.4} & \best{$4.13$\sd{0.70}} & \emph{$2.91$\sd{0.69} (ep.\ 78)} & \emph{$97.1\%$} \\
2D Gray--Scott ($\times 10^{-3}$) & $125.8$\sd{4.7} & \best{$5.22$\sd{0.37}} & \emph{$1.01$\sd{0.10} (ep.\ 19)} & \emph{$96.6\%$} \\
2D Kolmogorov NS ($\times 10^{-3}$) & $17.07$\sd{2.09} & \best{$4.35$\sd{0.42}} & \emph{$2.97$\sd{0.25} (ep.\ 22)} & \emph{$90.2\%$} \\
2D Guide-Field MHD ($\times 10^{-2}$) & $6.81$\sd{1.69} & \best{$3.38$\sd{1.01}} & \emph{$2.90$\sd{0.82} (ep.\ 7)} & \emph{$87.7\%$} \\
\bottomrule
\end{tabularx}
\end{table*}

\begin{table*}[t]
\centering
\caption{\textbf{2D Cahn--Hilliard from zero-output cold start ($h_0 = 0$, $G_{\theta_0} \equiv 0$, MSE $\times 10^{-4}$).} Validation ($784$ windows) and test ($784$ windows) MSE when all methods train from cold start rather than the shared offline-pretrained checkpoint $\theta_0$. Coarse Physics, EKI, Global/Local PPO, and the Neural Emulator Method all remain at the unparameterized coarse MSE.}
\label{tab:app-ch-coldstart}
\fontsize{8}{9.5}\selectfont
\renewcommand{\arraystretch}{1.08}
\setlength{\tabcolsep}{4pt}
\begin{tabular*}{\textwidth}{@{\extracolsep{\fill}}lccccccc@{}}
\toprule
Metric & Coarse / EKI / PPO / Emul. & ES & Closure-RL & Stage-II-Only & Stage-I-Only & \textbf{\ourmethod{} (Ours)} & \emph{Solver BPTT (oracle)} \\
\midrule
Validation MSE & $4.630$ & $4.624$ & $3.495$ & $2.136$ & \second{$1.719$} & \best{$1.704$} & \emph{$0.411$} \\
Test MSE & $4.520$ & $4.514$ & $3.424$ & $2.113$ & \second{$1.679$} & \best{$1.658$} & \emph{$0.421$} \\
\bottomrule
\end{tabular*}
\end{table*}

\subsection{Multi-Horizon Rollouts and Visualizations}
\label{app:visualizations}

\begin{table*}[t]
\centering
\caption{\textbf{3D POM per-variable 4-hour test trajectory MSE.} Overall and per-variable MSE across $(u, v, T, S, \eta)$ averaged over the $48$ test windows, normalized per field by the unparameterized coarse solver's 4-hour validation RMS error (Appendix~\ref{app:pom}).}
\label{tab:app-pom-test}
\fontsize{8}{9.5}\selectfont
\renewcommand{\arraystretch}{1.06}
\setlength{\tabcolsep}{7pt}
\begin{tabularx}{\textwidth}{@{}>{\raggedright\arraybackslash}Xcccccc@{}}
\toprule
Method & Total MSE & $u$ & $v$ & $T$ & $S$ & $\eta$ \\
\midrule
Coarse Physics & $1.1588$ & $1.0982$ & $1.1188$ & $1.1155$ & $1.4515$ & $1.0101$ \\
\midrule
EKI & $1.1588$ & $1.0982$ & $1.1188$ & $1.1155$ & $1.4515$ & $1.0101$ \\
PPO & $1.1581$ & $1.0970$ & $1.1168$ & $1.1153$ & $1.4513$ & $1.0102$ \\
ES & $1.1324$ & $1.0655$ & $1.0547$ & $1.0990$ & $1.4405$ & $1.0021$ \\
Closure-RL & $1.1178$ & $1.0631$ & $1.0275$ & $1.0764$ & $1.4161$ & $1.0057$ \\
Neural Emulator Method & \second{$0.6142$} & \second{$0.5539$} & \second{$0.5426$} & \second{$0.3577$} & \second{$0.7463$} & \second{$0.8703$} \\
\midrule
\textbf{\ourmethod{} (Ours)} & \best{$0.3466$} & \best{$0.0932$} & \best{$0.1082$} & \best{$0.1349$} & \best{$0.5694$} & \best{$0.8275$} \\
\emph{Solver BPTT (oracle)} & \emph{$0.3321$} & \emph{$0.1073$} & \emph{$0.1119$} & \emph{$0.1756$} & \emph{$0.5667$} & \emph{$0.6988$} \\
\bottomrule
\end{tabularx}
\end{table*}

\begin{table*}[t]
\centering
\caption{\textbf{2D out-of-window cumulative validation MSE at $80$ and $160$ steps.} Rollouts extend to $4\times$ and $8\times$ the $20$-step training horizon across Cahn--Hilliard, Kolmogorov NS, and guide-field MHD.}
\label{tab:app-2d-long}
\fontsize{8}{9.5}\selectfont
\renewcommand{\arraystretch}{1.06}
\setlength{\tabcolsep}{7pt}
\begin{tabularx}{\textwidth}{@{}>{\raggedright\arraybackslash}Xccccc@{}}
\toprule
Benchmark & Horizon & Coarse Physics & EKI & \textbf{\ourmethod{} (Ours)} & \emph{Solver BPTT (oracle)} \\
\midrule
2D Cahn--Hilliard ($\times 10^{-4}$) & 80 steps ($4\times$) & $13.20$ & $7.63$ & \best{$2.97$} & \emph{$2.00$} \\
2D Cahn--Hilliard ($\times 10^{-4}$) & 160 steps ($8\times$) & $27.39$ & $33.11$ & \best{$12.45$} & \emph{$10.64$} \\
\addlinespace[2pt]
2D Kolmogorov NS ($\times 10^{-2}$) & 80 steps ($4\times$) & $6.22$ & $1.89$ & \best{$1.35$} & \emph{$1.01$} \\
2D Kolmogorov NS ($\times 10^{-2}$) & 160 steps ($8\times$) & $15.54$ & $6.34$ & \best{$4.97$} & \emph{$3.67$} \\
\addlinespace[2pt]
2D Guide-Field MHD ($\times 10^{-2}$) & 80 steps ($4\times$) & $23.94$ & $11.71$ & \best{$9.76$} & \emph{$8.28$} \\
2D Guide-Field MHD ($\times 10^{-2}$) & 160 steps ($8\times$) & $39.14$ & $19.08$ & \best{$15.40$} & \emph{$13.02$} \\
\bottomrule
\end{tabularx}
\end{table*}

\begin{figure*}[tp]
  \centering
  \makeatletter\if@twocolumn\figfile[width=\textwidth]{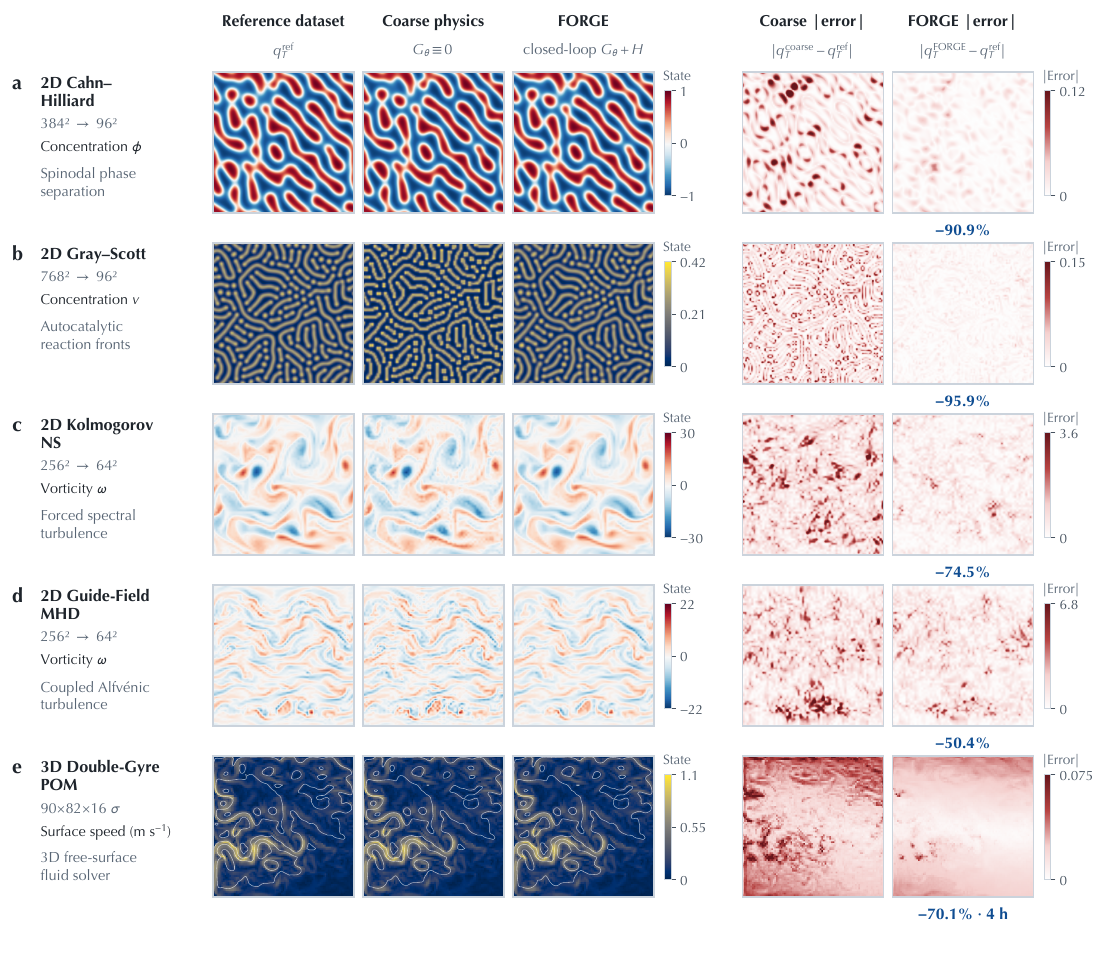}\else\figfile[width=0.92\textwidth]{FigS0_five_benchmarks_fields.pdf}\fi\makeatother
  \caption{\textbf{Full five-benchmark terminal-state comparison across all 2D and 3D regimes.} Side-by-side coarse-grained reference fields ($q_T^{\mathrm{ref}}$), unparameterized Coarse Physics ($G_\theta \equiv 0$), closed-loop \ourmethod{}, and matched-scale pointwise absolute errors ($|q_T - q_T^{\mathrm{ref}}|$) at step~$20$ for 2D Cahn--Hilliard (\textbf{a}), Gray--Scott (\textbf{b}), Kolmogorov NS (\textbf{c}), and guide-field MHD (\textbf{d}), and at $4\text{ h}$ for 3D Double-Gyre POM (\textbf{e}, upper-layer speed with sea-surface height contours and normalized multi-variable error). Blue annotations report test-set MSE reductions relative to Coarse Physics (Tables~\ref{tab:main-2d}--\ref{tab:main-3d}).}
  \label{fig:app-five-regimes}
\end{figure*}

\begin{figure*}[tp]
  \centering
  \makeatletter\if@twocolumn\figfile[width=\textwidth]{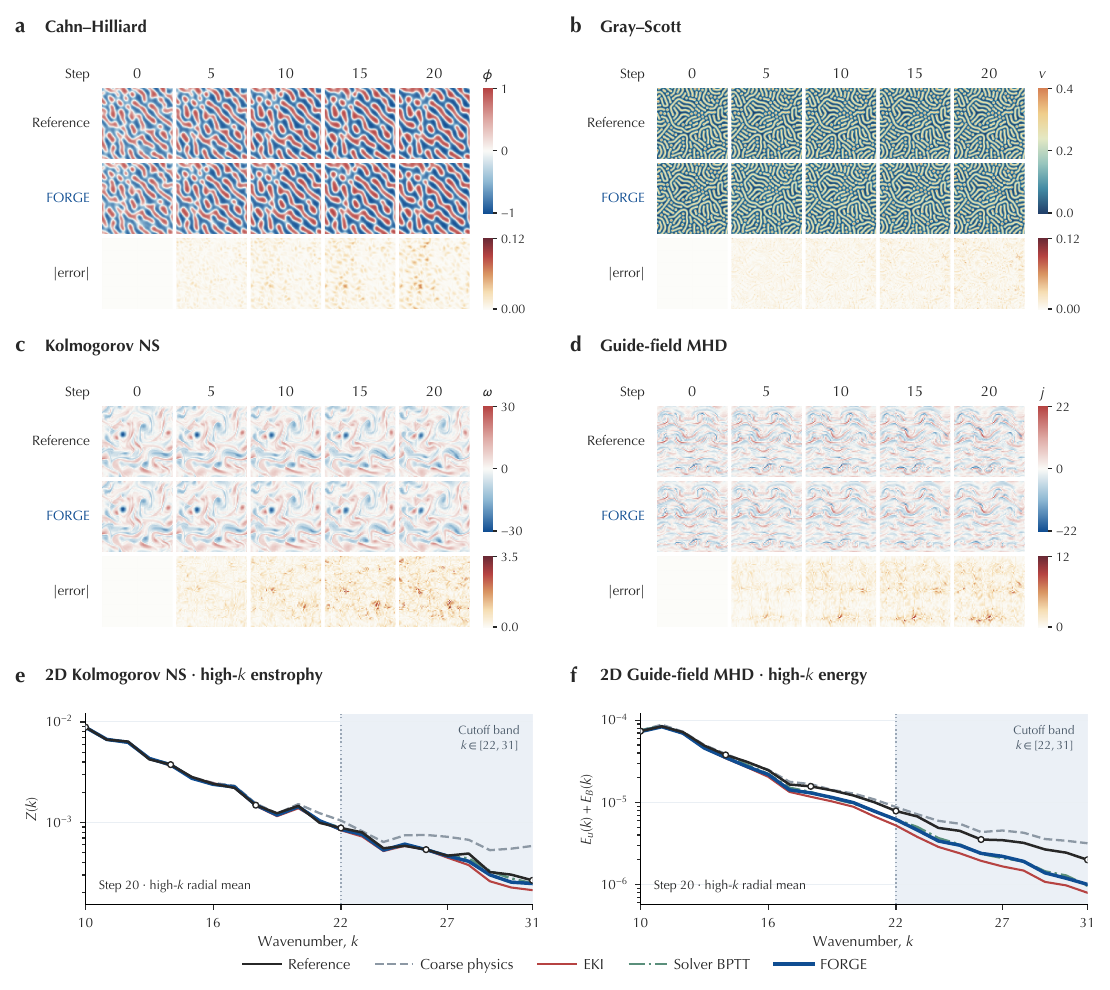}\else\figfile[width=0.88\textwidth]{FigS1_extended_2d_rollouts_spectra.pdf}\fi\makeatother
  \caption{\textbf{Spatiotemporal rollout evolution and high-wavenumber terminal spectra on the four 2D benchmarks.} \textbf{a--d}, Rollout snapshots across steps $0, 5, 10, 15, 20$ showing coarse-grained reference fields (top), \ourmethod{} rollouts (middle), and pointwise absolute errors $|q_t - q_t^{\mathrm{ref}}|$ (bottom) for Cahn--Hilliard concentration $\phi$ (\textbf{a}), Gray--Scott reactant $v$ (\textbf{b}), Kolmogorov vorticity $\omega$ (\textbf{c}), and guide-field MHD current density $j = \Delta\psi_m$ (\textbf{d}). \textbf{e, f}, High-wavenumber ($k \in [10, 31]$) step-20 radial enstrophy spectrum $Z(k)$ in Kolmogorov NS (\textbf{e}) and total hydromagnetic energy spectrum $E_u(k)+E_B(k)$ in guide-field MHD (\textbf{f}), with the grid cutoff band ($k \in [22, 31]$) shaded.}
  \label{fig:app-2d-rollouts}
\end{figure*}

\paragraph{2D rollouts and high-wavenumber spectra.}
Fig.~\ref{fig:app-five-regimes} compares terminal states and pointwise errors across all five benchmarks, Fig.~\ref{fig:app-2d-rollouts} tracks $0$--$20$ step 2D evolution and high-wavenumber spectra, and Table~\ref{tab:app-2d-long} summarizes $80$- and $160$-step 2D validation errors ($688/288$ windows on Cahn--Hilliard; $72/8$ on Kolmogorov NS and MHD). On Cahn--Hilliard and Gray--Scott, \ourmethod{} preserves sharp phase boundaries and reaction fronts where Coarse Physics blurs interfaces (Fig.~\ref{fig:app-five-regimes}a,b; Fig.~\ref{fig:app-2d-rollouts}a,b) and avoids EKI's late-time coarsening drift on Cahn--Hilliard (Table~\ref{tab:app-2d-long}). On Kolmogorov NS and guide-field MHD, \ourmethod{} suppresses the coarse-grid energy pile-up in the cutoff band ($k \in [22, 31]$, Fig.~\ref{fig:app-2d-rollouts}e,f) and remains stable through $160$ steps (Table~\ref{tab:app-2d-long}).

\begin{figure*}[tp]
  \centering
  \makeatletter\if@twocolumn\figfile[width=0.96\textwidth]{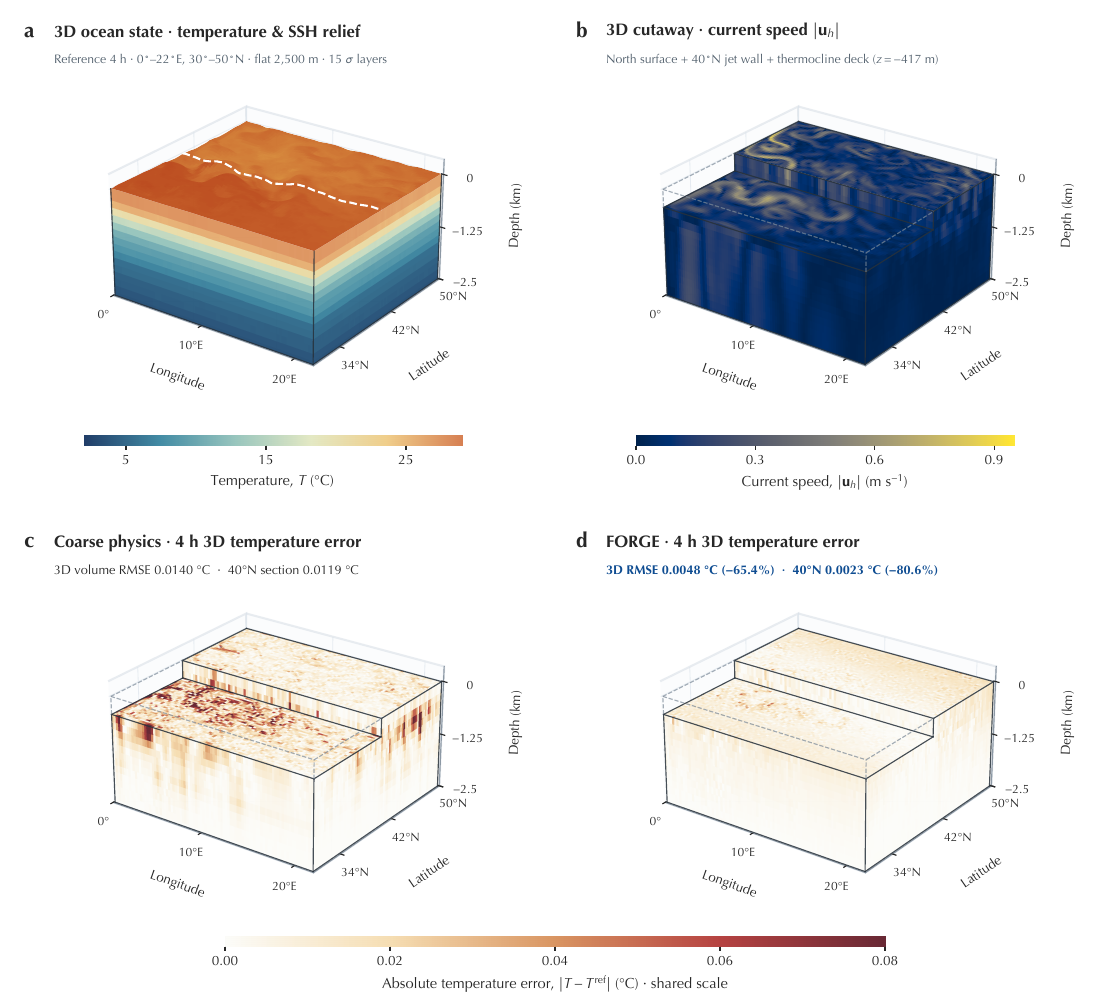}\else\figfile[width=0.90\textwidth]{FigS2_ocean_3d_transects.pdf}\fi\makeatother
  \caption{\textbf{3D POM volumetric state, cutaway circulation, and 4-hour 3D temperature error (validation member~$6$, day~$7{,}000$).} \textbf{a}, Full 3D ocean temperature block on the closed rectangular basin ($0^\circ$--$22^\circ\text{E}$, $30^\circ$--$50^\circ\text{N}$, flat depth $H_{\mathrm{b}} = 2{,}500\text{ m}$, $15$ uniform $\sigma$-layers) with vertically exaggerated sea-surface height ($\eta$) relief and the $40^\circ\text{N}$ transect trace (dashed). \textbf{b}, 3D step-cutaway of reference horizontal current speed $|\mathbf{u}_h|$ exposing the northern sea surface, the $40^\circ\text{N}$ jet wall, and the horizontal thermocline deck ($z = -417\text{ m}$). \textbf{c, d}, 4-hour 3D absolute temperature error $|T - T^{\mathrm{ref}}|$ on the same cutaway block and shared scale ($0$--$0.08\,^\circ\text{C}$) for Coarse Physics (\textbf{c}, 3D volume $\text{RMSE} = 0.0140\,^\circ\text{C}$; $40^\circ\text{N}$ section $0.0119\,^\circ\text{C}$) and \ourmethod{} (\textbf{d}, 3D $\text{RMSE} = 0.0048\,^\circ\text{C}$, $-65.4\%$; $40^\circ\text{N}$ section $0.0023\,^\circ\text{C}$, $-80.6\%$).}
  \label{fig:app-ocean}
\end{figure*}

\begin{figure*}[tp]
  \centering
  \figfile[width=0.88\textwidth]{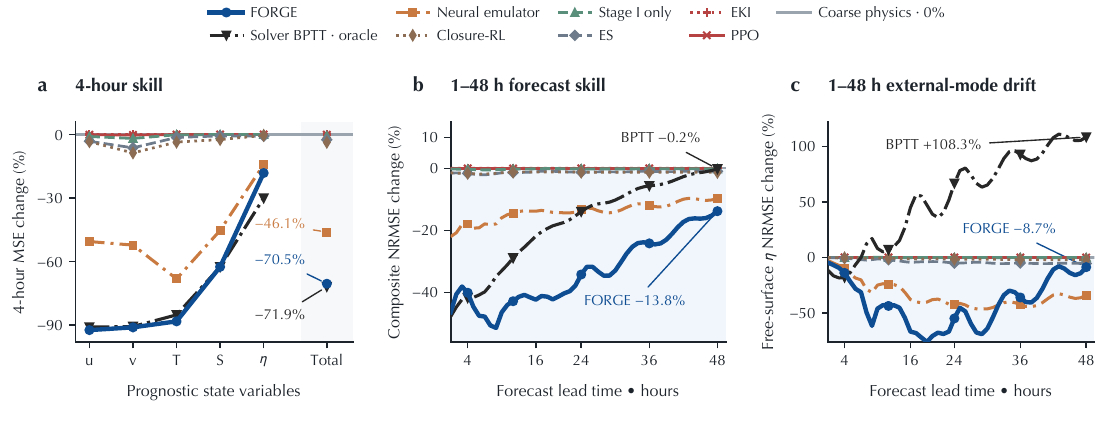}
  \caption{\textbf{3D POM per-variable validation skill and 48-hour forecast stability.} \textbf{a}, Percentage change in 4-hour validation MSE relative to Coarse Physics ($16$ validation windows) across $(u,v,T,S,\eta)$ and Total. \textbf{b}, Out-of-window composite NRMSE change relative to Coarse Physics across eight continuous $1$--$48\text{ h}$ validation rollouts ($12\times$ the training window). \textbf{c}, Out-of-window free-surface elevation $\eta$ NRMSE change over $1$--$48\text{ h}$, reaching $-8.7\%$ for \ourmethod{}, $-33.9\%$ for the Neural Emulator Method, and $+108.3\%$ for 300-epoch Solver BPTT at $48\text{ h}$.}
  \label{fig:app-ocean-mechanics}
\end{figure*}

\paragraph{3D ocean circulation and 48-hour stability.}
In 3D POM (Table~\ref{tab:app-pom-test} and Fig.~\ref{fig:app-ocean}), coarse truncation error concentrates along the western-boundary jet and upper thermocline. Across the $48$ test windows (Table~\ref{tab:app-pom-test}), \ourmethod{} reduces 4-hour velocity and temperature MSE by $88\%$--$92\%$ (outperforming 300-epoch Solver BPTT on $u, v, T$) and salinity MSE by $61\%$; on the 3D cutaway showcase (validation member~$6$, day~$7{,}000$; Fig.~\ref{fig:app-ocean}c,d), it suppresses upper-thermocline and $40^\circ\text{N}$ jet temperature errors across the basin. Because the conservation projector in \eqref{eq:app-pom-discrete} acts only on baroclinic tendencies, closures affect free-surface elevation $\eta$ indirectly through baroclinic pressure gradients. Over 48-hour rollouts (Fig.~\ref{fig:app-ocean-mechanics}), \ourmethod{} achieves the best composite skill while keeping $\eta$ error below Coarse Physics, whereas Solver BPTT overfits the 4-hour window and exhibits large free-surface drift.

\subsection{Response Accuracy and Tangent Verification}
\label{app:closed-loop-check}

\begin{table*}[t]
\centering
\caption{\textbf{Response prediction accuracy on unseen output perturbations across all five benchmarks.} Disjoint held-out perturbations $\delta a^{\mathrm{val}}$ are applied from the same initial states to compare $\mathcal{R}\,\delta a^{\mathrm{val}}$ (or Neural Emulator rollouts) against true solver differences $d^{\mathrm{val}}$ via median relative response error ($\mathrm{RRE} = \|d^{\mathrm{val}} - \mathcal{R}\,\delta a^{\mathrm{val}}\|_2 / \|d^{\mathrm{val}}\|_2$) and median cosine similarity across active Stage~II rounds.}
\label{tab:app-response}
\fontsize{8}{9.5}\selectfont
\renewcommand{\arraystretch}{1.10}
\setlength{\tabcolsep}{7pt}
\begin{tabularx}{\textwidth}{@{}>{\raggedright\arraybackslash}lXcc@{}}
\toprule
Benchmark & Operator / Surrogate & Median RRE ($\downarrow$) & Median Cosine ($\uparrow$) \\
\midrule
2D Cahn--Hilliard & \ourmethod{} (Stage~II) & $0.203$ & $0.979$ \\
2D Gray--Scott & \ourmethod{} (Stage~II) & $0.465$ & $0.885$ \\
2D Kolmogorov NS & \ourmethod{} (Stage~II) & $0.440$ & $0.898$ \\
2D Guide-Field MHD & \ourmethod{} (Stage~II) & $0.432$ & $0.902$ \\
3D POM & \ourmethod{} (Stage~II, 4h) & $0.267$ & $0.966$ \\
\midrule
2D Kolmogorov NS & Stage~II w/o warm-up ($h_0 = 0$) & $0.674$ & $0.751$ \\
\midrule
2D Cahn--Hilliard & Neural Emulator Method ($100$ closure calls) & $3.721$ & $0.163$ \\
2D Gray--Scott & Neural Emulator Method ($480$ closure calls) & $1.056$ & $0.487$ \\
3D POM & Neural Emulator Method ($4$ hourly steps) & $0.243$ & $0.970$ \\
\bottomrule
\end{tabularx}
\end{table*}

\paragraph{Response operator versus solver emulators.}
On unseen output perturbations (Table~\ref{tab:app-response}), Stage~II's causal operator $\mathcal{R}$ achieves $0.89$--$0.98$ median cosine similarity across all five benchmarks. By contrast, unrolling autoregressive solver emulators over $100$--$480$ closure calls in 2D accumulates large drift ($\mathrm{RRE} > 1$). On 4-step 3D POM, the emulator matches short-horizon forward differences closely, yet differentiating through its unconstrained network without damped inverse regularization \eqref{eq:adjoint-cg} leaves closed-loop test MSE nearly twice that of \ourmethod{} (Table~\ref{tab:app-pom-test}).

\paragraph{Why Stage~I warm-up matters.}
Stage~II fits a locally linear response around the current rollout. Starting directly from cold start ($h_0 = 0$) in strongly nonlinear regimes degrades local response fidelity: on Kolmogorov NS, skipping Stage~I raises $\mathrm{RRE}$ from $0.440$ to $0.674$ and lowers cosine similarity from $0.898$ to $0.751$ (Table~\ref{tab:app-response}), explaining why Stage~I warm-up stabilizes subsequent full-network updates (Table~\ref{tab:ablation}).

\paragraph{Tangent alignment with Solver BPTT.}
Across $20$ rounds on 2D Kolmogorov NS, we compare Stage~II's predicted trajectory shift $\mathcal{R}\,\delta a(\delta\theta)$ and directional loss change against the exact Solver BPTT tangent $\mathcal{J}_a\,\delta a(\delta\theta)$ and true gradient inner product for each full-network update $\delta\theta = \theta_{\mathrm{proj}} - \theta^{\mathrm{I}}$. The predicted and exact tangents achieve $0.914$ median cosine similarity, agree in descent sign on all $20$ rounds, and match directional-derivative magnitude within $28\%$ median relative error.

\clearpage
{\small\raggedright
\setlength{\bibsep}{3.0pt plus 0.8pt minus 0.5pt}
\putbib[refs]
}
\end{bibunit}

\end{document}